\documentclass{article}

\usepackage[final]{corl_2026}

\usepackage{graphicx}
\usepackage{booktabs}
\usepackage{multirow}
\usepackage{amsmath}
\usepackage{capt-of}

\newcommand{\figref}[1]{Fig.~\ref{#1}}

\newcommand{\tabref}[1]{Table~\ref{#1}}

\title{Functional-SLAM: Interaction-Aware Mapping with Online Functional Scene Graphs}

\author{
  \textbf{Xinggang Hu$^{1,2}$, Chenyangguang Zhang$^{3}$, Zihan Zhu$^{3}$}\\
  \textbf{Ruida Zhang$^{1}$, Xiangkui Zhang$^{2}$, Xiangyang Ji$^{1,\dagger}$}\\
  $^{1}$Tsinghua University \qquad
  $^{2}$Dalian University of Technology \qquad
  $^{3}$ETH Zurich\\
  $\dagger$ Corresponding author
}

\begin{document}
\maketitle

\begin{abstract}
Existing SLAM systems lack modeling of the functional relations required for fine-grained robotic interaction. Functional 3D scene graphs can represent relations between objects and interaction elements, but existing methods rely on offline reconstruction, making them inadequate for real-time interaction in real-world exploration. 
To address this limitation, we propose Functional-SLAM, the first framework that continuously and recursively maintains a functional scene graph as an online SLAM state.
The framework combines anchor-keyframe geometry with functional-context constraints for persistent node maintenance, accumulates multi-frame evidence through temporal relations to commit stable functional edges, and supplements visual loop-closure candidates with functional topology in scenes with repetitive appearance or degraded texture.
Experiments show that Functional-SLAM efficiently constructs stable functional maps online, substantially improving runtime over offline methods while maintaining highly competitive accuracy. Compared with peer SLAM systems, it further improves pose estimation accuracy through functional-topology-assisted loop closure.
The code is publicly available at \url{https://github.com/Hbelief1998/Functional-SLAM-CoRL_2026}.

\end{abstract}


\section{Introduction}
	
With the development of robotics, augmented reality, and embodied intelligence, it has become increasingly important to construct scene representations that bridge 3D scene perception and downstream interactive tasks. 
Existing SLAM systems can already estimate camera motion and build geometric maps in real time~\cite{izadi2011kinectfusion,mur2017orb,campos2021orb,engel2017direct,murai2025mast3r}. 
Semantic SLAM~\cite{mccormac2017semanticfusion,tian2022kimera,zhu2024sni}, object-level SLAM~\cite{yang2019cubeslam,nicholson2018quadricslam,wu2023object,li2025dqo}, and open-vocabulary 3D mapping~\cite{huang2023visual,yamazaki2024open,peng2023openscene,kerr2023lerf,qin2024langsplat} further lift maps to the object and semantic levels. 
However, for fine-grained interaction, robots need not only to recognize objects, but also to understand the operability of interaction elements and their functional effects on objects. Existing SLAM systems generally lack modeling of these small interaction elements and their functional relations, preventing them from performing fine-grained tasks that require functional reasoning and manipulation planning, such as ``picking up the kettle to pour some water.''

Functional 3D scene graphs provide a more suitable representation for interactive perception~\cite{zhang2025open,rotondi2025fungraph,werby2025keysg,fu2026funfact,hu2026hierarchical}.However, existing methods rely on known poses, depth inputs, or offline reconstruction, and typically construct functional graphs offline on stable geometry. In real deployment, robots inevitably explore unknown environments; such an offline paradigm incurs high latency and deployment cost, failing to meet the needs of immediate interaction. Therefore, we advance functional graphs from offline construction to an online map state continuously updated during SLAM.

Under this online setting, functional graph construction faces two key challenges. 
First, functional observations rely on continuously updated camera poses and local geometry, while online poses may drift during front-end tracking and are continuously refined by back-end optimization. This issue is particularly pronounced for small interactive elements such as handles, knobs, and buttons, because they are small in scale, visually similar, and weakly supported by geometry; even slight pose or reconstruction errors may lead to incorrect associations. 
Second, functional edges exhibit strong single-frame ambiguity: occlusion, local overlap, and competition among neighboring candidates can cause ambiguous object affiliations for interaction elements, which need to be gradually stabilized through temporal evidence. 

To this end, we propose Functional-SLAM, the first framework that continuously maintains a functional scene graph as an online SLAM state. The system treats open-vocabulary perception outputs as frame-wise functional observations. To address node maintenance challenges caused by online pose drift, pose updates, and cross-frame confusion of small interaction elements, it maintains node geometry in anchor-keyframe local coordinates so that it remains synchronized with pose optimization, and preserves persistent functional nodes using geometric, semantic, and functional-relation cues. To resolve single-frame relation ambiguity, it introduces temporal relation posteriors and commits functional edges only when multi-frame evidence becomes stable. Furthermore, the online functional graph is distilled into structured map memory, providing functional-topological candidates for visual loop closure in scenes with repetitive appearance or degraded texture. 
Experiments show that Functional-SLAM yields temporally stable functional mapping online, gaining competitive accuracy and much faster speed than prior offline functional scene graph methods. 
Compared with SLAM counterparts, it further improves pose estimation accuracy through functional-topology-assisted loop-closure candidate generation.

The main contributions of this paper are as follows:
\begin{itemize}
    \item The Functional-SLAM framework, which, to the best of our knowledge, is the first to advance functional scene graphs from offline construction to an online mapping state during SLAM, achieving superior performance in localization and functional graph construction.

    \item An online functional graph maintenance scheme that integrates anchor-keyframe geometry, functional-context-constrained node association, and temporal relation posteriors to enable online updates of stable functional nodes and edges.

    \item A functional-topology-assisted loop-closure mechanism that leverages stable nodes and functional topology to supplement visual loop-closure candidates in scenes with repetitive appearance or degraded texture, further improving pose estimation accuracy.
\end{itemize}

    
    
\section{Related Work}

\noindent\textbf{Online Pose Estimation and Mapping.}
Classical visual SLAM/VO methods perform online camera tracking and mapping through feature matching, direct photometric consistency, or sparse/semi-dense optimization~\cite{klein2007parallel,mur2015orb,mur2017orb,campos2021orb,engel2014lsd,engel2017direct,forster2014svo}. Dense SLAM has further evolved from traditional geometric fusion~\cite{newcombe2011dtam,newcombe2011kinectfusion,whelan2012kintinuous,whelan2015elasticfusion,dai2017bundlefusion,runz2017co,schops2019bad} to online mapping with neural implicit representations~\cite{mildenhall2021nerf,sucar2021imap,zhu2022nice,zhu2024nicer,johari2023eslam,yang2022vox,sandstrom2023point,wang2023co,zhang2023go} and 3D Gaussians~\cite{kerbl20233d,matsuki2024gaussian,keetha2024splatam,sandstrom2025splat,yan2024gs,yugay2023gaussian,huang2024photo,ha2024rgbd}, producing higher-fidelity geometry and appearance maps. Recent learning-based methods improve pose estimation and long-sequence 3D reconstruction using dense matching, reconstruction priors, or 3D foundation models~\cite{teed2021droid,teed2023deep,murai2025mast3r,maggio2026vggt,chen2025ttt3r,deng2025vggt}. Meanwhile, semantic SLAM~\cite{mccormac2017semanticfusion,tian2022kimera,zhu2024sni,zhu2025semgauss,ji2024neds}, object-level SLAM~\cite{yang2019cubeslam,nicholson2018quadricslam,yang2019monocular,mccormac2018fusion++,runz2017co,sucar2020nodeslam,wu2023object,li2025dqo}, and open-vocabulary 3D mapping~\cite{huang2023visual,yamazaki2024open,peng2023openscene,kerr2023lerf,qin2024langsplat,gu2024conceptgraphs,takmaz2023openmask3d,wu2024opengaussian} extend maps beyond geometry to categories, instances, and language semantics. 
However, these methods mainly focus on camera trajectories, geometric structures, object instances, or semantic labels, while still lacking perception of operable elements and their functional relations, making them insufficient for fine-grained interaction tasks.

\noindent\textbf{Functional Scene Graphs.}
3D indoor scene understanding typically focuses on objects, instances, and open-vocabulary semantic perception~\cite{atzmon2018point,choy20194d,hu2021vmnet,hua2018pointwise,huang2023segment3d,landrieu2018large,li2018pointcnn,qi2017pointnet,qi2017pointnet++,thomas2019kpconv,weder2023alster,weder2024labelmaker,engelmann20203d,han2020occuseg}, while affordance understanding further predicts interactable regions or parts in a scene~\cite{banerjee2024introducing,chen2022alignsdf,cho2024dense,deng20213d,do2018affordancenet,fan2024hold,fang2018demo2vec,mo2022o2o,nagarajan2020learning,nagarajan2019grounded,xu2022partafford,ye2022s,ye2024g,yoshida2024text,zhai2022one,zhang2024ddf,zhang2024moho}. Functional 3D scene graphs extend these representations by introducing operable elements and their functional relations with objects, enabling finer-grained interaction reasoning~\cite{armeni20193d,koch2024lang3dsg,rosinol20203d,rosinol2021kimera,takmaz2025search3d,wald2020learning,wu2021scenegraphfusion,wu2023incremental,xu2017scene,yang2018graph,zhang2021exploiting,li2021ifr,koch2024open3dsg,gu2024conceptgraphs}. 
OpenFunGraph~\cite{zhang2025open} first defines functional 3D scene graphs as open-vocabulary graph structures composed of objects, interaction elements, and functional relations, and constructs them from posed RGB-D sequences using VLMs and LLMs.
Subsequent works~\cite{rotondi2025fungraph,werby2025keysg,fu2026funfact,hu2026hierarchical} further extend this direction through functional-element modeling, hierarchical structures, probabilistic relation reasoning, and task-relevant scene graphs.
However, existing methods often rely on known poses or offline reconstruction, and thus struggle to meet the demands of real-time exploration.

\section{Method}
\subsection{Overview}
\begin{figure}[t]
  \centering
   \includegraphics[width=1.0\linewidth]{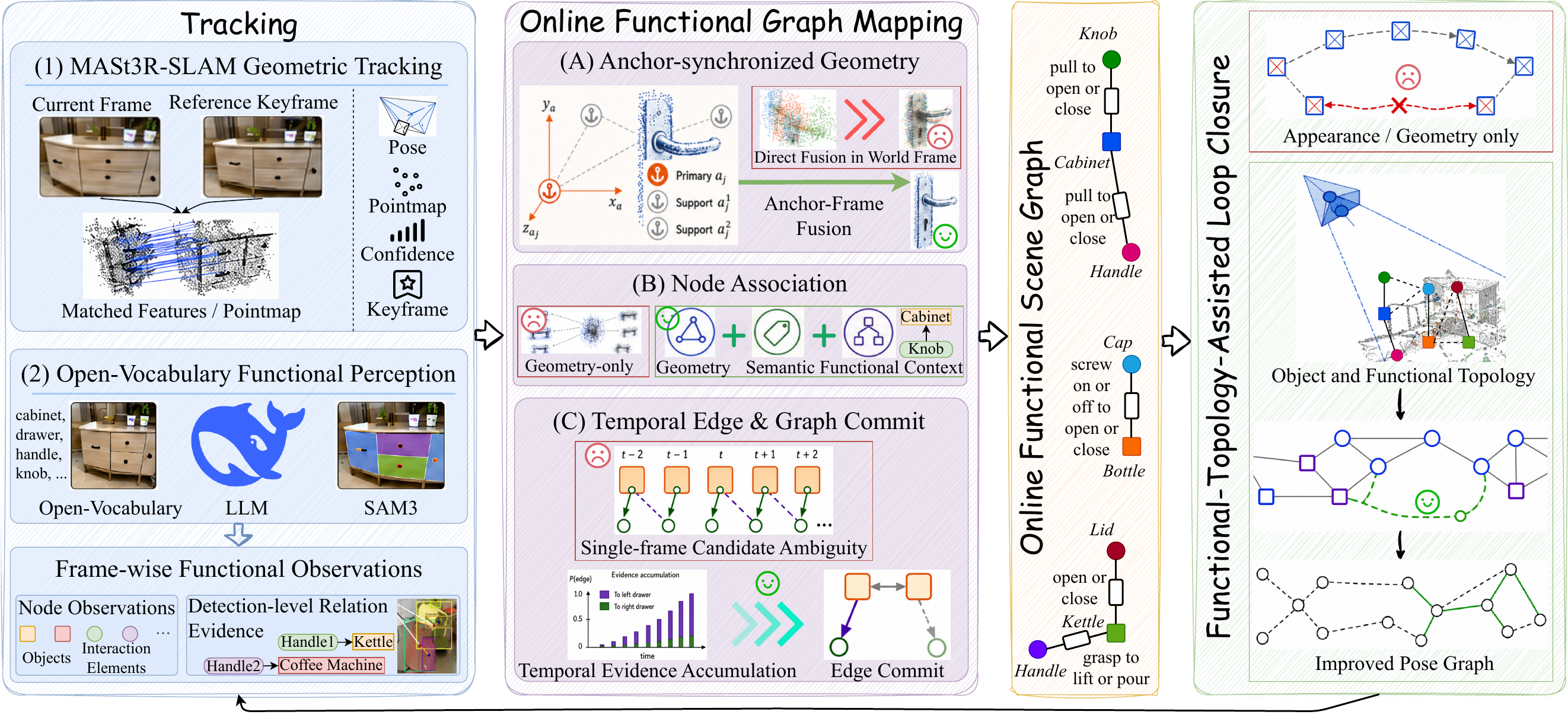}
    \vspace{-6mm} 
   \caption{Functional-SLAM pipeline.
The system performs tracking, online functional graph mapping, and functional-topology-assisted loop closure.
   }
   \label{fig:teaser}
    \vspace{-6mm} 
\end{figure}

{We summarize our pipeline in Fig.~\ref{fig:teaser}. Given a continuous RGB image stream $\{I_t\}_{t=1}^{N}$, our system simultaneously performs camera tracking, online functional-graph mapping, and loop closure detection. Following OpenFunGraph~\cite{zhang2025open}, the online functional graph is represented as $G_t=(U_t,O_t,R_t)$, where $U_t$ denotes robot-operable interaction elements, $O_t$ scene objects, and $R_t$ their functional relations. Incoming frames are handled by a frontend tracking module that uses MASt3R-SLAM~\cite{murai2025mast3r} as the geometric backbone for SLAM-state estimation and keyframe selection, while open-vocabulary perception produces frame-level functional observations. In parallel, the mapping module recursively fuses these observations into a persistent graph under the latest optimized poses. Loop closure combines visual retrieval with object-level and functional-topology candidates from the online functional graph to maintain global consistency.}

\subsection{Tracking}

Functional-SLAM uses MASt3R-SLAM's tracking front end as the geometric backbone and generates frame-level functional observations during tracking. Given the current image $I_t$, MASt3R-SLAM matches it with the reference keyframe, establishes inter-frame correspondences from MASt3R-predicted dense pointmaps and matched features, and optimizes the current-frame pose using the matched pointmaps. Functional-SLAM keeps this geometric tracking objective unchanged, while reusing the estimated pose, canonical pointmap, confidence map, and keyframe decision to support 3D lifting of functional observations and subsequent online mapping.

Alongside geometric tracking, the system constructs open-vocabulary functional observations for the current frame. It uses image-level open-vocabulary labels and DeepSeek-based reasoning to obtain scene-relevant functional perception targets, which are then localized and segmented by SAM3. For each detection, the system records its label $\ell_i^t$, role $r_i^t$, 2D box $b_i^t$, segmentation mask $m_i^t$, detection confidence $s_i^t$, and back-projects the masked region into 3D support $P_i^t$ using the current-frame pointmap and pose. This yields the node observation set $N_t=\{n_i^t\}$, where $r_i^t\in\{O,U\}$ denotes object and interaction-element observations, respectively.

Based on these node observations, the system further constructs detection-level functional relation evidence. Let $n_o^t\in N_t$ be an object observation and $n_u^t\in N_t$ an interaction-element observation. Their relation observation in the current frame is written as $(n_o^t\leftarrow n_u^t,\eta_{ou}^t)$, where $\eta_{ou}^t$ measures the support for ``interaction element $n_u^t$ acts on object $n_o^t$'':
\begin{equation}
\small
\eta_{ou}^t
=
\delta_{ou}^t
\min(s_o^t,s_u^t)
\frac{|m_u^t\cap m_o^t|}{|m_u^t|}.
\end{equation}
Here, $\delta_{ou}^t\in\{0,1\}$ is a functional-structure gate, which equals 1 when their current functional contexts are compatible and 0 otherwise. $(s_o^t,s_u^t)$ and $(m_o^t,m_u^t)$ denote the confidences and masks of the object and interaction-element observations, respectively. Finally, the current-frame functional observation is denoted as $Z_t=(N_t,B_t)$, where $B_t=\{(n_o^t\leftarrow n_u^t,\eta_{ou}^t)\}$. These observations remain local, noisy detection-level evidence, which will be lifted to persistent-node-level relational evidence through node association and temporally stabilized in the mapping stage.

\subsection{Online Functional Graph Mapping}

The single-frame functional observation $Z_t=(N_t,B_t)$ obtained in tracking cannot be directly written into the map. On the one hand, SLAM front-end drift and back-end optimization can make nodes fixed in the world frame geometrically inconsistent, easily causing association errors for small, weakly supported, and visually similar interaction elements. On the other hand, candidate competition caused by occlusion and local overlap leads to severe affiliation ambiguity in single-frame functional relations.
To address these issues, the mapping stage recursively fuses frame-level observations into a persistent functional graph $G_t$ under the latest optimized poses. For node association, it combines anchor-keyframe geometry with functional-context constraints, maintaining nodes in local coordinates to overcome pose-induced misalignment, while using geometric, semantic, and functional-relation cues to stably match current observations $N_t$ to the persistent node set $V_{t-1}$. For single-frame relation ambiguity, it accumulates detection-level observations $B_t$ into multi-frame evidence, thereby effectively eliminating incorrect affiliations.

\subsubsection{Node Association}

Offline functional scene graphs are typically built on stable poses from SfM or offline reconstruction, allowing node geometry to be written once into fixed 3D coordinates. In online SLAM, however, front-end tracking accumulates drift, while back-end optimization continuously refines historical keyframe poses, causing the world positions, image projections, and relative spatial relations of functional nodes to change over time. Unlike object-level SLAM, which mainly focuses on large object instances, Functional-SLAM must maintain finer-grained interaction elements such as handles, knobs, and buttons. These elements are usually small, visually similar, and weakly supported by 3D geometry, making cross-frame association more prone to failure under pose drift. Under this online setting, node maintenance must address both geometry synchronization and identity association. Anchor-keyframe-based pose-synchronized geometry provides persistent nodes with geometry readout updated according to the latest SLAM state, avoiding projection and comparison in stale world coordinates. Functional-context-constrained node association further combines geometric consistency, open-vocabulary labels, and local relational evidence to stably associate current observations $N_t$ with the persistent node set $V_{t-1}$.

\noindent\textbf{Anchor-Keyframe-Based Geometry.}
For each stable functional node $v_j$, Functional-SLAM binds an anchor keyframe $a_j$ and maintains the node's canonical geometry in the local coordinate system of this keyframe, rather than fixing it as static world-space geometry. 
Anchor construction has two stages: candidate selection and stable promotion. 
Let $H_j^t$ denote the observation history associated with node $v_j$ up to time $t$, and $A_j^t$ the set of candidate keyframes corresponding to these observations. 
For each candidate anchor keyframe $a\in A_j^t$, the system ranks it by single-frame observation quality:
\begin{equation}
\small
a_j^\star
=
\arg\max_{a\in A_j^t} Q_{\mathrm{view}}(a),
\qquad
Q_{\mathrm{view}}(a)
=
(b_a,q_a,g_a)_{\mathrm{lex}},
\qquad
q_a
=
s_a\bar{c}_a\frac{|m_a|}{|I_a|},
\quad
g_a
=
|P_a|.
\end{equation}
Here, $a_j^\star$ is the current best candidate anchor keyframe, and $(\cdot)_{\mathrm{lex}}$ denotes lexicographic comparison. $b_a$ indicates whether the observation avoids image-boundary truncation, $s_a$ is the detection confidence, $\bar{c}_a$ is the average geometric confidence of valid 3D points inside the mask, $|m_a|/|I_a|$ is the visible area ratio, and $P_a$ is the valid point set obtained by back-projecting the masked region. 

A candidate anchor keyframe is stabilized only after cross-frame support and world-geometry verification. For later observations, the system transforms their 3D support into the candidate-frame coordinates, verifies consistency with the provisional fused geometry in center, scale, and bounding region, and checks agreement with the node's historical world geometry. During projection, association, and relation reasoning, the node's global geometry is dynamically recovered from the latest anchor-frame pose, keeping it synchronized with back-end optimization.



\noindent\textbf{Functional-Context-Constrained Node Association.}
Anchor-keyframe-synchronized geometry provides node predictions under the current SLAM state. Based on these predictions, node association establishes a one-to-one persistent mapping for current observations and provides endpoint correspondences for subsequent relation posterior construction. Given the current observation set $N_t$ and persistent node set $V_{t-1}$, the system performs role-wise matching, where objects and interaction elements are matched only within their own categories. For an observation $n_i^t\in N_t$ and a node $v_j\in V_{t-1}$, the matching score is defined as
\begin{equation}
\small
S_{ij}^t
=
\phi_{\mathrm{geo}}(n_i^t,v_j)
+
\phi_{\mathrm{sem}}(n_i^t,v_j)
+
\phi_{\mathrm{ctx}}(n_i^t,v_j,B_t).
\end{equation}
Here, $\phi_{\mathrm{geo}}$ measures consistency in projection overlap, 3D center distance, and scale-adaptive gating based on node-predicted geometry under the latest optimized poses; $\phi_{\mathrm{sem}}$ measures consistency between the open-vocabulary label and the node's historical semantic statistics; and $\phi_{\mathrm{ctx}}$ uses the current relation evidence $B_t$ to constrain functional affiliation and penalizes candidate associations that conflict with stable historical affiliations. Details are provided in the supplementary material.

The system then uses the Hungarian algorithm to solve the optimal one-to-one matching over candidate associations. Matched observations update the corresponding node states, while unmatched ones are initialized as new candidate nodes and stabilized through verification in subsequent frames.

\subsubsection{Temporal Functional-Edge Posterior and Graph Commitment}

Node association only determines the persistent identities of relation endpoints. During online SLAM mapping, single-frame relation evidence is highly prone to affiliation ambiguity due to viewpoint occlusion and local overlap. For example, interaction elements such as handles on neighboring objects are often occluded or overlapped from certain viewpoints, making their affiliations difficult to distinguish. Therefore, the system lifts single-frame relation observations to persistent evidence.

Let $\pi_t(n)$ denote the persistent node corresponding to the observed node $n$ in the current frame. For an object node $v_o$ and an interaction-element node $v_u$, their current-frame relation support $u_t$ is determined by the highest-confidence observation mapped to this node pair:
\begin{equation}
\small
u_t(v_o\leftarrow v_u)
=
\max \left\{
\eta_{ou}^t
\mid
(n_o^t\leftarrow n_u^t,\eta_{ou}^t)\in B_t,\ 
\pi_t(n_o^t)=v_o,\ 
\pi_t(n_u^t)=v_u
\right\}.
\end{equation}
The system then accumulates historical valid support to obtain the repeated multi-frame support $r_t$, and selects the currently most reliable object $v_o^\star$ from the associated object set $O(v_u)$:
\begin{equation}
\small
r_t(v_o\leftarrow v_u)
=
\sum_{s\le t} u_s(v_o\leftarrow v_u),
\qquad
v_o^\star
=
\arg\max_{v_o\in O(v_u)}
r_t(v_o\leftarrow v_u).
\end{equation}
The relation $v_o^\star\leftarrow v_u$ is committed to the online functional graph only when it has sufficient multi-frame support and consistent recent best-object selections. Each interaction element retains only its most reliable relation, ensuring unique affiliation and avoiding conflicts.

\subsection{Functional-Topology-Assisted Loop Closure}
In parallel with local tracking and mapping, we perform loop closure to inject long-range constraints into the factor graph and suppress drift. MASt3R-SLAM detects loop candidates via visual retrieval and validates them through geometric verification, but visual retrieval alone misses valid loops under repetitive appearance, degraded texture, or large viewpoint changes. Our Functional-SLAM additionally exploits the online functional graph $G_t$ as a structured map memory that proposes object-level and functional-topology loop closure candidates. These are merged with MASt3R-SLAM's visual candidates and jointly undergo geometric verification and loop closure optimization.

Let $k$ be the query keyframe and $k'$ a historical candidate. We design two scores for loop closure candidates:
$S_G(k, k')$, capturing \emph{object-level topology}, and
$S_{FT}(k, k')$, capturing \emph{functional topology}.
A keyframe pair $(k, k')$ is retained as a candidate if
\begin{equation}
\small
    S_G(k, k') > \tau_G \quad \text{or} \quad S_{FT}(k, k') > \tau_{FT},
\end{equation}
and among the surviving pairs, we keep the top-$K$.


Specifically, for each keyframe $k$ we construct a local functional-topological signature $\Gamma_k$ from $G_t$ as follows. For every visible stable node $v_i$, we form a graphlet $g_i=\{v_i,v_{i_1},v_{i_2}\}$ together with its two nearest stable neighbors, and define $\Gamma_k=\{g_i\}$. 
Given $g\in\Gamma_k$ and $h\in\Gamma_{k'}$, $S_{\mathrm{sem}}(g,h)$ denotes the semantic compatibility ratio between the two graphlets under the best node correspondence. Two corresponding nodes are considered compatible if they share the same persistent node ID, or have the same role and open-vocabulary label. Shape consistency $S_{\mathrm{shape}}(g,h)$ is computed from the distance between normalized edge-length vectors. Please refer to the supplementary material for details.
Let $w(g,h)$ denote the topological matching score between graphlets $(g,h)$. 
For each query graphlet $g$, let
$h_g=\arg\max_{h\in\Gamma_{k'}}w(g,h)$
be the best-matched graphlet. The object-level topological similarity is defined as
\begin{equation}
\small
S_G(k,k') = \frac{1}{|\Gamma_k|} \sum_{g \in \Gamma_k} w(g, h_g),
\quad \text{with} \quad
w(g, h) = S_{\mathrm{sem}}(g, h)\, S_{\mathrm{shape}}(g, h).
\end{equation}
We further introduce functional-topology candidates. Let $\psi(g, h_g)$ denote the fraction of committed functional relations in graphlet $g$ that are preserved by its match $h_g$ under the induced node correspondence. The functional-topological score is defined as
\begin{equation}
\small
S_{\mathrm{FT}}(k,k')
=
S_G(k,k')\bigl(1+S_F(k,k')\bigr),
\qquad
S_F(k,k')
=
\frac{
\sum_{g\in\Gamma_k} w(g,h_g)\psi(g,h_g)
}{
\sum_{g\in\Gamma_k} w(g,h_g)
},
\end{equation}
where $S_F(k,k')$ is the functional-relation preservation rate weighted by the graphlet matching quality $w(g, h_g)$. Relative to $S_G$, which captures purely object-level topological similarity, $S_{\mathrm{FT}}$ additionally rewards functional-relation consistency under the established local correspondences.

\section{Experiments}

\subsection{Experimental Setup}


\noindent\textbf{Baselines.}
For localization, we compare with diverse pose estimation and mapping baselines, including classical visual SLAM/VO methods DSO\cite{engel2017direct}, ORB-SLAM3~\cite{campos2021orb}, and DPVO~\cite{teed2023deep}, the object-level SLAM method DQO-MAP~\cite{li2025dqo}, and learning-based pose estimation or 3D reconstruction methods TTT3R~\cite{chen2025ttt3r}, DROID-SLAM~\cite{teed2021droid}, VGGT-SLAM~\cite{maggio2026vggt}, VGGT-SLAM-2.0~\cite{maggio2026vggt2}, VGGT-Long~\cite{deng2025vggt}, and MASt3R-SLAM~\cite{murai2025mast3r}. For functional scene graph construction, we compare with offline functional graph methods OpenFunGraph~\cite{zhang2025open}, FunGraph~\cite{rotondi2025fungraph}, and KeySG~\cite{werby2025keysg}. For fairness in functional reasoning, all LLM-based functional graph methods use the same DeepSeek-V4-Flash as ours.
All experiments are conducted on a single NVIDIA RTX 3090 GPU.

\noindent\textbf{Datasets and preprocessing.}
We evaluate on SceneFun3D~\cite{delitzas2024scenefun3d} and FunGraph3D~\cite{zhang2025open}, both of which contain real indoor scenes with annotations related to functional scene graphs, making them suitable for evaluating object, interaction-element, and functional-relation mapping. 
Since the raw data contain many segments irrelevant to functional scene graphs, and some sequences suffer from large inter-frame gaps, textureless regions, and severe motion blur, the original sequences are not directly suitable for SLAM. We therefore select 18 sequences from each dataset and crop continuous segments containing valid objects, interaction elements, and functional relations. 

\noindent\textbf{Metrics.}
For localization, we report ATE RMSE to measure the absolute error between the estimated and reference trajectories~\cite{grupp2017evo}. For functional scene graph construction, we follow the Recall@K protocol of OpenFunGraph~\cite{zhang2025open} and evaluate the recall of nodes, and functional-relation triplets. Specifically, we report node R@3/R@10 and triplet R@5/R@10 to measure the overall quality of node recognition, node association, and functional-edge prediction in the online functional graph.

\subsection{Localization}
\begin{figure}[h]
\centering

\newlength{\locLeftWidth}
\newlength{\locRightWidth}
\newlength{\locVGap}
\newlength{\locTargetHeight}
\newsavebox{\ateCurveBox}
\newsavebox{\trajBox}

\setlength{\locLeftWidth}{0.33\linewidth}
\setlength{\locRightWidth}{0.66\linewidth}
\setlength{\locVGap}{1.5mm}

\sbox{\ateCurveBox}{%
    \includegraphics[width=\locLeftWidth]{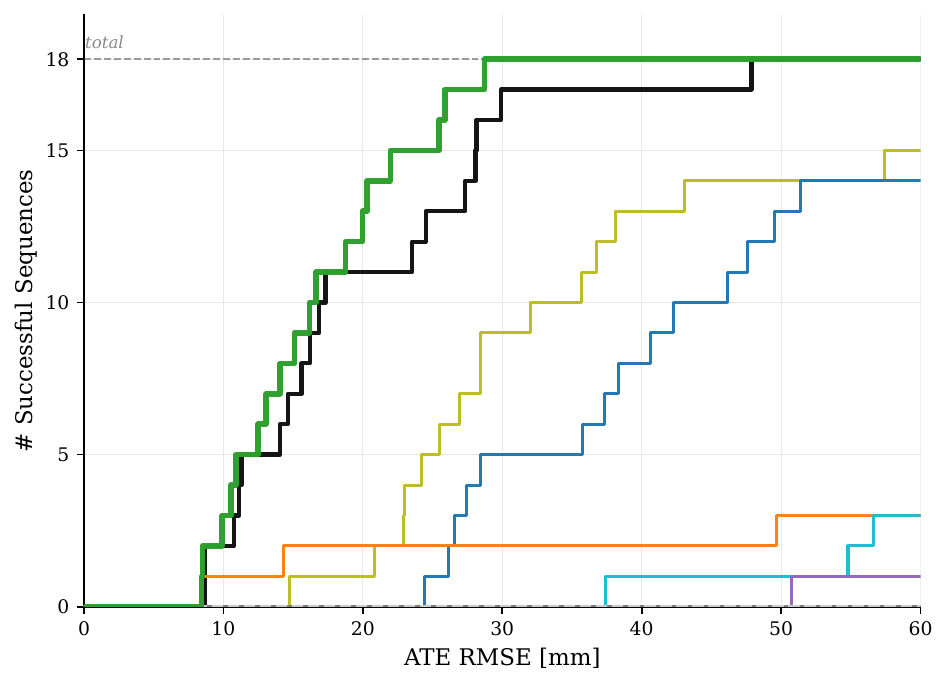}%
}
\sbox{\trajBox}{%
    \includegraphics[width=\locLeftWidth]{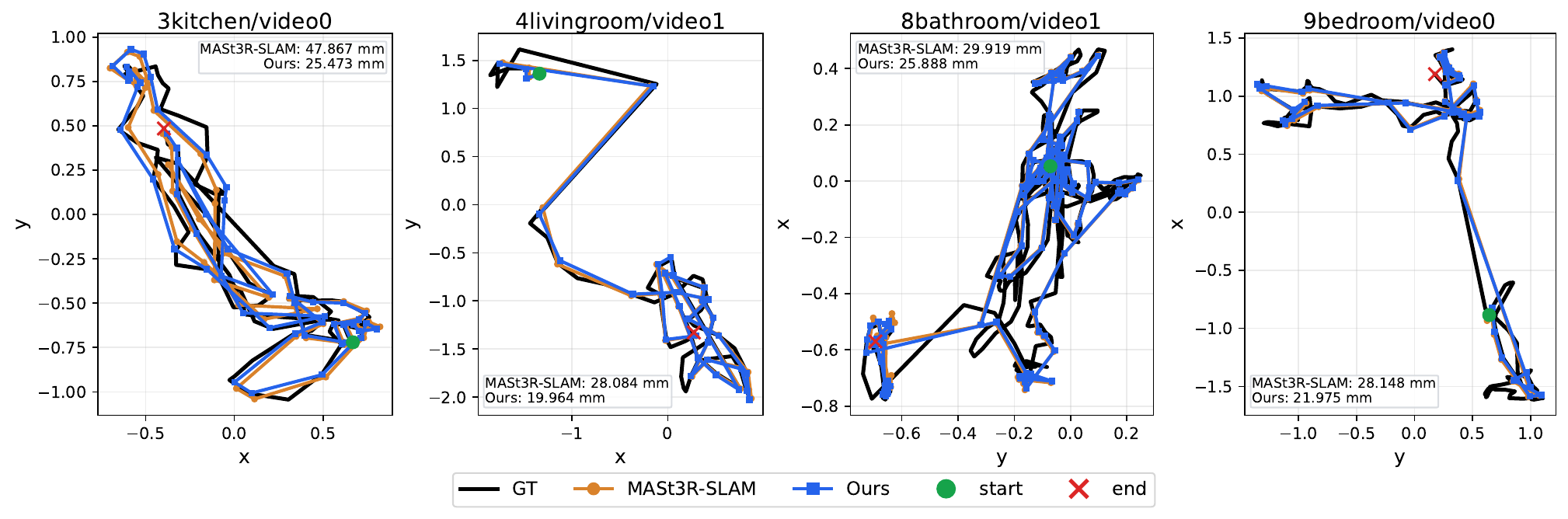}%
}

\setlength{\locTargetHeight}{%
    \dimexpr
    \ht\ateCurveBox+\dp\ateCurveBox+
    \ht\trajBox+\dp\trajBox+
    \locVGap
    \relax
}

\begin{minipage}[t]{\locLeftWidth}
    \centering
    \vspace{0pt}
    \usebox{\ateCurveBox}

    \vspace{\locVGap}

    \usebox{\trajBox}
\end{minipage}
\hfill
\begin{minipage}[t]{\locRightWidth}
    \centering
    \vspace{0pt}
    \includegraphics[
        width=\linewidth,
        height=\locTargetHeight
    ]{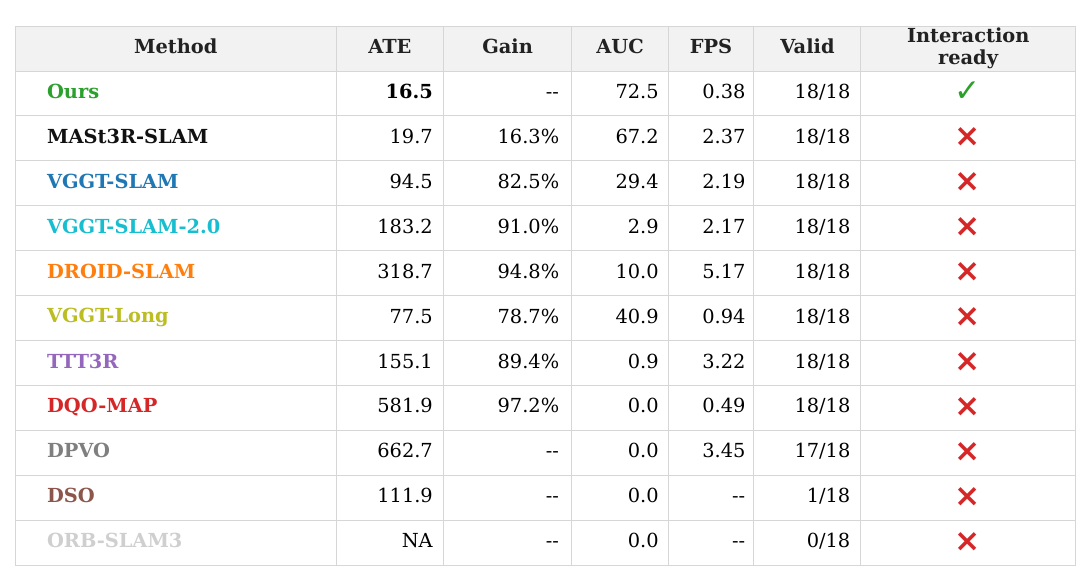}
\end{minipage}

\caption{Localization evaluation. Top-left: cumulative success curves under 0--60 mm ATE RMSE thresholds. Bottom-left: representative trajectories showing drift correction by functional-topology-assisted loop closure. Right: summary of ATE RMSE, AUC, runtime, and valid trajectories.}
\label{fig:loc}
\vspace{-3mm}

\end{figure}

As shown in \figref{fig:loc}, although the evaluated sequences still contain extreme challenges such as abrupt viewpoint changes, weak texture, and motion blur, causing traditional SLAM methods to fail in most cases, Functional-SLAM achieves strong localization results.
On FunGraph3D, our method reduces ATE RMSE by 16.3\% over the strongest baseline, MASt3R-SLAM, achieving the best accuracy. This improvement mainly comes from functional-topology-assisted loop closure: stable topological cues complement loop-closure candidates missed by purely visual retrieval under repetitive appearance or degraded texture, effectively correcting accumulated drift, as shown by the bottom-left trajectory in \figref{fig:loc}.
On SceneFun3D, since the sequences are texture-rich and contain very few loop closures, visual cues are already sufficient for recall. Our method therefore remains comparable to the baseline, with an average ATE RMSE of 33.25~mm. This shows that when visual features are sufficient, our method preserves the localization performance of the geometric backbone.

\subsection{Functional Scene Graph}
To decouple the influence of pose errors on functional reasoning, we evaluate mapping performance under both GT pose and Ours pose settings.

As shown in \tabref{tab:node_evaluation} and \tabref{tab:triplet_evaluation}, Functional-SLAM shows clear advantages on FunGraph3D. This dataset contains richer objects and interaction elements, with relatively smooth camera motion. Benefiting from Anchor-Keyframe-Based Geometry, functional-context-constrained node association, and temporal relation posteriors, the system constructs more accurate and stable online functional graphs.
On SceneFun3D, significant pose drift limits our performance under Ours pose, since fine-grained elements are highly sensitive to pose and 3D alignment errors. As a result, our method does not outperform some offline baselines that rely on GT pose. Nevertheless, under the same Ours pose setting, Functional-SLAM still substantially outperforms other methods, showing that the proposed online maintenance mechanism effectively mitigates node confusion caused by pose noise.

\begin{figure}[h]
  \centering
   \includegraphics[width=1.0\linewidth]{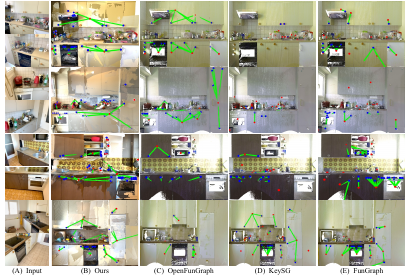}
    \vspace{-6mm} 
   \caption{Qualitative comparison of functional scene graphs. 
   }
   \label{fig:fun_graph} 
\end{figure}

\figref{fig:fun_graph} presents qualitative results in several challenging scenes. Functional-SLAM constructs more complete functional nodes and more accurate functional relations, especially for small interaction elements and complex object-affiliation relations. Note that the other methods in the figure are shown under the GT pose setting. When directly using Ours pose, their functional graphs become highly noisy; the corresponding results are provided in the appendix.

\begin{table*}[t]
\centering
\caption{Node evaluation on the SceneFun3D~\cite{delitzas2024scenefun3d} and FunGraph3D~\cite{zhang2025open} datasets.}
\label{tab:node_evaluation}
\scriptsize
\setlength{\tabcolsep}{3.2pt}
\begin{tabular}{llcccccccccccc}
\toprule
\multirow{3}{*}{Pose setting} & \multirow{3}{*}{Method}
& \multicolumn{6}{c}{SceneFun3D}
& \multicolumn{6}{c}{FunGraph3D} \\
\cmidrule(lr){3-8} \cmidrule(lr){9-14}
& & \multicolumn{2}{c}{Objects}
& \multicolumn{2}{c}{Inter. Elements}
& \multicolumn{2}{c}{Overall Nodes}
& \multicolumn{2}{c}{Objects}
& \multicolumn{2}{c}{Inter. Elements}
& \multicolumn{2}{c}{Overall Nodes} \\
& & R@3 & R@10 & R@3 & R@10 & R@3 & R@10
& R@3 & R@10 & R@3 & R@10 & R@3 & R@10 \\
\midrule
\multirow{3}{*}{GT pose}
& OpenFunGraph~\cite{zhang2025open} & \textbf{43.48} & \textbf{48.91} & 52.86 & 56.39 & 50.16 & 54.23 & \textbf{49.01} & \textbf{54.97} & 45.45 & 49.82 & 46.71 & 51.64 \\
& FunGraph~\cite{rotondi2025fungraph}     & 35.87 & 41.30 & \textbf{66.08} & \textbf{68.72} & \textbf{57.37} & \textbf{60.82} & 47.02 & 51.66 & \textbf{54.91} & \textbf{56.00} & \textbf{52.11} & \textbf{54.46} \\
& KeySG~\cite{werby2025keysg}        & \textbf{43.48} & 47.83 & 47.14 & 47.14 & 46.08 & 47.34 & 32.45 & 42.38 & 38.91 & 39.64 & 36.62 & 40.61 \\
\midrule
\multirow{4}{*}{Ours pose}
& OpenFunGraph~\cite{zhang2025open} & 29.35 & 31.52 & 14.98 & \textbf{16.74} & 19.12 & 21.00 & 44.37 & 49.67 & 28.36 & 33.09 & 34.04 & 38.97 \\
& FunGraph~\cite{rotondi2025fungraph}     & 29.35 & 31.52 & 15.86 & \textbf{16.74} & 19.75 & 21.00 & 47.02 & 51.66 & 35.27 & 37.09 & 39.44 & 42.25 \\
& KeySG~\cite{werby2025keysg}        & 30.43 & 31.52 &  9.25 & 10.13 & 15.36 & 16.30 & 30.46 & 40.40 & 22.18 & 23.27 & 25.12 & 29.34 \\
& Ours         & \textbf{47.83} & \textbf{51.09} & \textbf{16.30} & 16.30 & \textbf{25.39} & \textbf{26.33} & \textbf{63.58} & \textbf{68.87} & \textbf{65.09} & \textbf{66.18} & \textbf{64.55} & \textbf{67.14} \\
\bottomrule
\end{tabular}
\end{table*}
\begin{table*}[t]
\centering
\caption{Triplet evaluation on the SceneFun3D~\cite{delitzas2024scenefun3d} and FunGraph3D~\cite{zhang2025open} datasets.}
\label{tab:triplet_evaluation}
\scriptsize
\setlength{\tabcolsep}{3.2pt}
\begin{tabular}{llcccccccccccc}
\toprule
\multirow{3}{*}{Pose setting} & \multirow{3}{*}{Method}
& \multicolumn{6}{c}{SceneFun3D}
& \multicolumn{6}{c}{FunGraph3D} \\
\cmidrule(lr){3-8} \cmidrule(lr){9-14}
& & \multicolumn{2}{c}{Node Assoc.}
& \multicolumn{2}{c}{Edge Pred.}
& \multicolumn{2}{c}{Overall Triplets}
& \multicolumn{2}{c}{Node Assoc.}
& \multicolumn{2}{c}{Edge Pred.}
& \multicolumn{2}{c}{Overall Triplets} \\
& & R@5 & R@10 & R@5 & R@10 & R@5 & R@10
& R@5 & R@10 & R@5 & R@10 & R@5 & R@10 \\
\midrule
\multirow{3}{*}{GT pose}
& OpenFunGraph~\cite{zhang2025open} & 27.19 & 30.70 & 53.23 & \textbf{91.43} & 14.47 & \textbf{28.07} & 36.99 & 41.10 & 48.15 & 93.33 & 17.81 & 38.36 \\
& FunGraph~\cite{rotondi2025fungraph}     & 18.42 & 19.74 & \textbf{85.71} & 86.67 & \textbf{15.79} & 17.11 & \textbf{41.78} & \textbf{42.12} & \textbf{96.72} & \textbf{96.75} & \textbf{40.41} & \textbf{40.75} \\
& KeySG~\cite{werby2025keysg}        & \textbf{35.09} & \textbf{35.53} & 43.75 & 43.21 & 15.35 & 15.35 & 19.52 & 23.29 & 66.67 & 61.76 & 13.01 & 14.38 \\
\midrule
\multirow{4}{*}{Our pose}
& OpenFunGraph~\cite{zhang2025open} & 7.02 & 8.33 & 50.00 & \textbf{89.47} & 3.51 & 7.46 & 23.97 & 28.08 & 48.57 & 91.46 & 11.64 & 25.68 \\
& FunGraph~\cite{rotondi2025fungraph}     & 9.65 & 10.09 & 72.73 & 73.91 & 7.02 & 7.46 & 27.74 & 28.42 & 97.53 & 97.59 & 27.05 & 27.74 \\
& KeySG~\cite{werby2025keysg}        & 4.39 & 4.39 & 30.00 & 30.00 & 1.32 & 1.32 & 4.79 & 8.22 & 92.86 & 87.50 & 4.45 & 7.19 \\
& Ours         & \textbf{11.40} & \textbf{11.84} & \textbf{84.62} & 85.19 & \textbf{9.65} & \textbf{10.09} & \textbf{42.12} & \textbf{44.86} & \textbf{97.56} & \textbf{97.71} & \textbf{41.10} & \textbf{43.84} \\
\bottomrule
\end{tabular}
\end{table*}

\begin{figure}[t]
\centering

\begin{minipage}[t]{0.525\linewidth}
    \centering
    \vspace{0pt}

    \captionof{table}{Ablation study on ATE, node evaluation, and triplet evaluation on the FunGraph3D~\cite{zhang2025open}.}
    \label{tab:ablation_study}
    \scriptsize
    \setlength{\tabcolsep}{2.5pt}
    \resizebox{\linewidth}{!}{%
    \begin{tabular}{@{}lccccc@{}}
    \toprule
    \multirow{2}{*}{Method}
    & ATE
    & \multicolumn{2}{c}{Overall Nodes}
    & \multicolumn{2}{c}{Overall Triplets} \\
    \cmidrule(lr){2-2} \cmidrule(lr){3-4} \cmidrule(lr){5-6}
    & RMSE & R@3 & R@10 & R@5 & R@10 \\
    \midrule
    w/o Node Stabilization        & 19.0 & 52.11 & 54.93 & 25.68 & 29.45 \\
    w/o Temporal Edge Posterior   & 17.9 & 64.08 & 66.67 & 31.51 & 35.27 \\
    w/o Functional Loop           & 19.7 & 61.74 & 64.32 & 35.96 & 38.36 \\
    \textbf{Ours}                          & \textbf{16.5} & \textbf{64.55} & \textbf{67.14} & \textbf{41.10} & \textbf{43.84} \\
    \bottomrule
    \end{tabular}
    }

    \vspace{1.5mm}

    \captionof{table}{Runtime comparison in FPS.}
    \label{tab:fps_comparison}
    \scriptsize
    \setlength{\tabcolsep}{4.0pt}
    \resizebox{\linewidth}{!}{%
    \begin{tabular}{@{}lcccc@{}}
    \toprule
    Method & OpenFunGraph & FunGraph & KeySG & \textbf{Ours} \\
    \midrule
    FPS & 0.02 & 0.06 & 0.05 & \textbf{0.38} \\
    \bottomrule
    \end{tabular}
    }

\end{minipage}
\hfill
\begin{minipage}[t]{0.447\linewidth}
    \centering
    \vspace{0pt}
    \includegraphics[width=\linewidth]{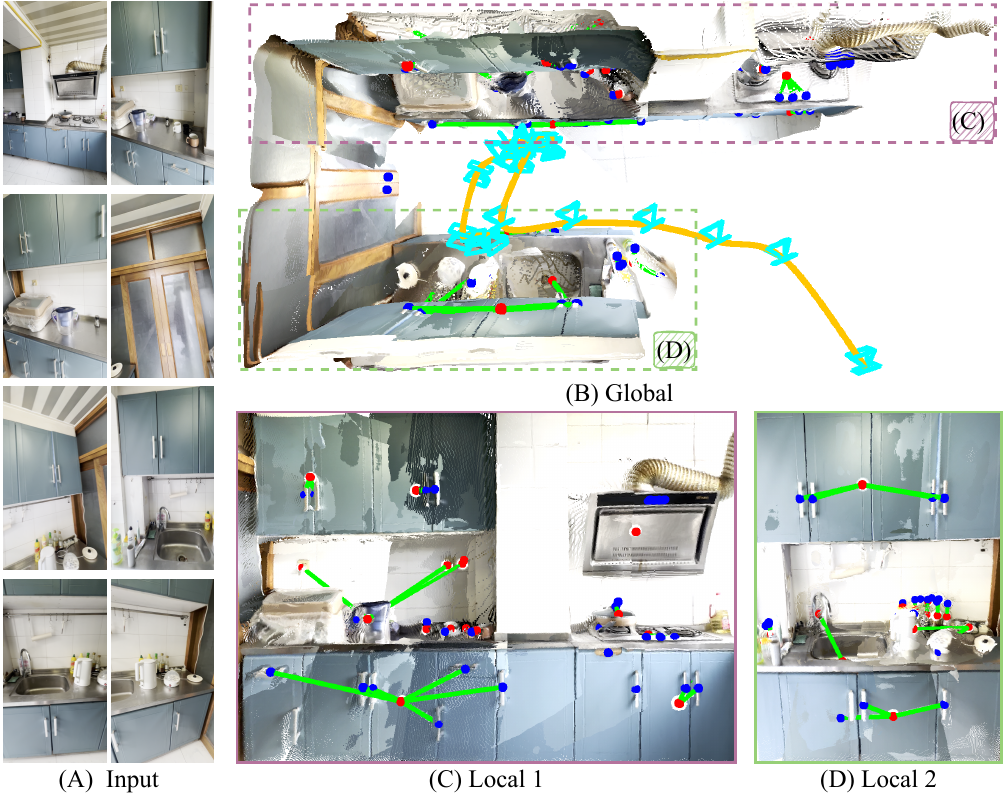}
    \vspace{-6mm}
    \captionof{figure}{Real-world scenes.}
    \label{fig:real}
\end{minipage}

\vspace{-3mm}
\end{figure}

\subsection{Ablation Study}

We evaluate three key components of Functional-SLAM. \textbf{w/o Node Stabilization} maintains node geometry in world coordinates and uses OpenFunGraph-style association. \textbf{w/o Temporal Edge Posterior} directly commits the highest-confidence relation from the current frame. \textbf{w/o Functional Loop} disables functional-topology-assisted loop closure.
As shown in \tabref{tab:ablation_study}, removing node stabilization degrades node association under pose drift, resulting in higher ATE and lower triplet recall. Removing temporal edge posteriors makes functional relations vulnerable to occlusion and local ambiguity, lowering triplet accuracy. Removing functional loop closure reduces loop-retrieval robustness in repetitive or low-texture scenes, degrading localization and functional graph quality.


\subsection{Real-World Scenes}
To evaluate real-world generalization, we test indoor sequences captured with a handheld iPhone in an intrinsics-free monocular mode. As shown in \figref{fig:real}, despite extreme challenges such as dense tiny elements, severe local occlusion, and sparse observations, the system successfully performs camera tracking and online functional mapping. \figref{fig:real}(B) shows the global trajectory, while \figref{fig:real}(C) and \figref{fig:real}(D) show local details, demonstrating strong practical applicability.


\subsection{Runtime Performance}
As shown in \figref{fig:loc}, online pose estimation baselines run slightly faster than our method, but achieve lower localization accuracy and, more importantly, cannot construct functional graphs for fine-grained interaction. 
Meanwhile, \tabref{tab:fps_comparison} shows our method's clear efficiency advantage over offline functional graph methods.
Its runtime can be further improved by performing functional observation and construction only on keyframes.
See Appendix for detailed runtime analysis.

\section{Conclusion}

We present Functional-SLAM, which, to the best of our knowledge, is the first framework that continuously maintains a functional scene graph as an online map state during SLAM. The system recursively fuses open-vocabulary perception results into persistent functional nodes and stable functional edges, and further leverages functional topology to assist loop-closure candidate generation. Experiments show that Functional-SLAM constructs temporally stable functional maps online, achieves higher efficiency than offline functional graph methods, and delivers competitive performance in both localization and functional graph construction.

\noindent\textbf{Limitations.}
Functional-SLAM depends on reliable visual tracking and open-vocabulary perception. In highly challenging scenarios with large inter-frame gaps, textureless regions, or severe motion blur, both pose estimation and functional scene graph construction may degrade.

\acknowledgments{This work was supported by the National Science and Technology Major Project 2025ZD1606303, Fundamental and Interdisciplinary Disciplines Breakthrough Plan of the Ministry of Education of China JYB2025XDXM503 and National Natural Science Foundation of China 62495092.}

\bibliography{main}

\clearpage
\appendix

\begin{center}
    {\Large\bfseries Supplementary Material}\\[0.5em]
    {\large Functional-SLAM: Interaction-Aware Mapping with Online Functional Scene Graphs}
\end{center}

\vspace{1em}

\section{Additional Qualitative Results}

\begin{figure}[t]
  \centering
   \includegraphics[width=1.0\linewidth]{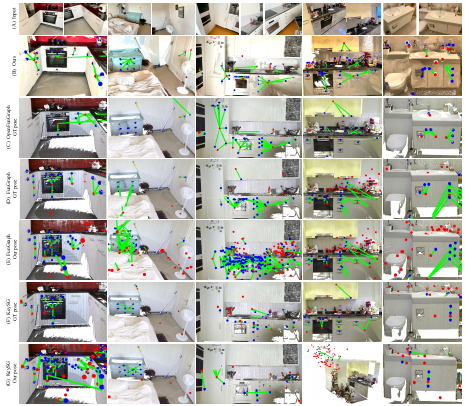}
   \caption{Qualitative comparison of functional scene graphs. 
   }
   \label{fig:sup_fun_graph} 
\end{figure}

Fig.~\ref{fig:sup_fun_graph} shows additional qualitative results of functional graph construction. Each column corresponds to one test scene. The first row shows the input image, while the following rows show the functional graphs constructed by Functional-SLAM and offline functional graph methods under different pose settings. Functional-SLAM constructs more complete and more stable functional scene graphs using online estimated poses. In contrast, although OpenFunGraph~\cite{zhang2025open}, FunGraph~\cite{rotondi2025fungraph}, and KeySG~\cite{werby2025keysg} can generate a certain number of functional nodes and edges under the GT-pose setting, their results are still more prone to missed detections, incorrect associations, or incomplete relations. Under the Functional-SLAM-pose setting, namely the Ours-pose setting, the degradation of these offline methods becomes more pronounced.

This difference mainly comes from the online functional graph maintenance mechanism. Functional-SLAM does not independently fuse functional observations from each frame into a fixed 3D reconstruction. Instead, it continuously maintains persistent functional nodes and relations during SLAM. The anchor-keyframe-based geometry keeps the geometric states of functional nodes synchronized with back-end pose optimization. The functional-context-constrained node association jointly uses geometric, semantic, and current relational evidence to maintain stable and accurate cross-frame node associations. The temporal relation posterior further suppresses erroneous functional edges caused by single-frame occlusion, overlap, and local ambiguity through multi-frame evidence. Therefore, even in the presence of pose drift or locally unstable geometry, Functional-SLAM can still obtain relatively clear and structurally consistent functional graphs.

It should be noted that OpenFunGraph often generates very few, or even no, nodes and edges under the Functional-SLAM-pose setting. This is mainly because the offline pipeline of OpenFunGraph relies on stable 3D fusion results to generate object-part candidates. When estimated poses are used, pose errors and drift degrade the fusion quality after projecting 2D masks back into 3D. For small parts such as handles, knobs, and buttons, such errors further reduce the valid 3D support, or make the parts no longer maintain stable overlap with their corresponding objects. As a result, the object-part candidate generation stage can hardly find valid candidates. Even if VLMs or LLMs are subsequently invoked, there is insufficient input for generating relational edges, leading to a large number of missing nodes and edges in the final functional graph.

The fusion strategies of FunGraph and KeySG are relatively less restrictive than that of OpenFunGraph. Therefore, they may still generate more nodes and edges under the Ours-pose setting. However, since these methods lack node geometry maintenance synchronized with online pose optimization and stable cross-frame association mechanisms, multi-frame observations under estimated poses are difficult to fuse correctly. Consequently, the number of functional nodes may increase, but their identities become confused; meanwhile, edge connections become dense but lack stable functional ownership. Such errors are especially likely to occur in regions with dense small parts or strong repetitive structures. The large number of cluttered nodes and edges produced by FunGraph and KeySG under the Ours-pose setting in the figure reflects the accumulated errors caused by the absence of an online stable maintenance mechanism.

Overall, the qualitative results further demonstrate that the advantages of Functional-SLAM are reflected not only in node or triplet recall, but also in the interpretability and temporal stability of the functional graph structure. By recursively maintaining the functional graph as an online SLAM state, Functional-SLAM can construct more accurate, more complete, and less noisy functional scene graphs under realistic estimated-pose conditions. In contrast, offline functional graph methods are more likely to suffer from missing nodes, missing edges, or confused relations when stable poses and online node maintenance mechanisms are unavailable.

\section{Runtime Optimization}

To improve online efficiency, we implement a runtime-optimized variant, denoted as Ours w/ RT. The full version performs open-vocabulary perception and online functional graph updates on every frame, which maximizes the density of functional observations. However, its major computational cost comes from semantic information acquisition modules such as RAM++~\cite{huang2025open}, DeepSeek~\cite{liu2024deepseek}, and SAM3~\cite{carion2025sam3segmentconcepts}. The key idea of the runtime-optimized version is that, once scene-level functional semantics become stable, functional perception and functional graph maintenance are performed only on keyframes, rather than repeatedly running the full pipeline on every regular tracking frame.

Specifically, the runtime optimization consists of two parts. First, for non-keyframes, the system skips RAM++, DeepSeek, and SAM3, and performs semantic information acquisition and functional observation generation only on keyframes. Meanwhile, non-keyframes also skip node association, edge association, and graph updates in OnlineFunctionalGraph. This avoids repeatedly executing time-consuming open-vocabulary reasoning and functional graph maintenance on a large number of non-keyframes. Second, we reduce the SAM3 processor resolution from 1008 to 672 to further decrease the cost of segmentation inference. Since functional graph construction relies on multi-frame key observations rather than dense semantic prediction on every frame, this downsampling strategy significantly improves runtime efficiency while preserving good functional graph quality.


\begin{table}[t]
\centering
\caption{Runtime comparison of major modules before and after runtime optimization on the 6kitchen/video0 sequence.}
\label{tab:rt_optimization_runtime}
\resizebox{\linewidth}{!}{
\begin{tabular}{lcccccc}
\toprule
Setting & FPS & Semantic Perception & Anchor Update & Node Assoc. & Edge Assoc. & FG Loop Retrieval \\
\midrule
Before RT & 0.377 & 2040.06 ms/frame & 60.90 ms/frame & 117.66 ms/frame & 12.76 ms/frame & 211.88 ms/frame \\
Ours w/ RT & 0.920 & 514.23 ms/frame & 53.77 ms/frame & 22.30 ms/frame & 4.76 ms/frame & 210.27 ms/frame \\
\bottomrule
\end{tabular}
}
\end{table}

\begin{table*}[t]
\centering
\caption{Node evaluation on the SceneFun3D~\cite{delitzas2024scenefun3d} and FunGraph3D~\cite{zhang2025open} datasets. Ours w/ RT denotes the real-time optimized variant. Best and second-best distinct results within each pose setting are highlighted in bold and underlined, respectively.}
\label{tab:node_evaluation_rt}
\scriptsize
\setlength{\tabcolsep}{3.2pt}
\begin{tabular}{llcccccccccccc}
\toprule
\multirow{3}{*}{Pose setting} & \multirow{3}{*}{Method}
& \multicolumn{6}{c}{SceneFun3D}
& \multicolumn{6}{c}{FunGraph3D} \\
\cmidrule(lr){3-8} \cmidrule(lr){9-14}
& & \multicolumn{2}{c}{Objects}
& \multicolumn{2}{c}{Inter. Elements}
& \multicolumn{2}{c}{Overall Nodes}
& \multicolumn{2}{c}{Objects}
& \multicolumn{2}{c}{Inter. Elements}
& \multicolumn{2}{c}{Overall Nodes} \\
& & R@3 & R@10 & R@3 & R@10 & R@3 & R@10
& R@3 & R@10 & R@3 & R@10 & R@3 & R@10 \\
\midrule
\multirow{3}{*}{GT pose}
& OpenFunGraph~\cite{zhang2025open} & \textbf{43.48} & \textbf{48.91} & \underline{52.86} & \underline{56.39} & \underline{50.16} & \underline{54.23} & \textbf{49.01} & \textbf{54.97} & \underline{45.45} & \underline{49.82} & \underline{46.71} & \underline{51.64} \\
& FunGraph~\cite{rotondi2025fungraph} & \underline{35.87} & 41.30 & \textbf{66.08} & \textbf{68.72} & \textbf{57.37} & \textbf{60.82} & \underline{47.02} & \underline{51.66} & \textbf{54.91} & \textbf{56.00} & \textbf{52.11} & \textbf{54.46} \\
& KeySG~\cite{werby2025keysg} & \textbf{43.48} & \underline{47.83} & 47.14 & 47.14 & 46.08 & 47.34 & 32.45 & 42.38 & 38.91 & 39.64 & 36.62 & 40.61 \\
\midrule
\multirow{5}{*}{Ours pose}
& OpenFunGraph~\cite{zhang2025open} & 29.35 & 31.52 & 14.98 & \textbf{16.74} & 19.12 & 21.00 & 44.37 & 49.67 & 28.36 & 33.09 & 34.04 & 38.97 \\
& FunGraph~\cite{rotondi2025fungraph} & 29.35 & 31.52 & 15.86 & \textbf{16.74} & 19.75 & 21.00 & 47.02 & 51.66 & 35.27 & 37.09 & 39.44 & 42.25 \\
& KeySG~\cite{werby2025keysg} & 30.43 & 31.52 & 9.25 & 10.13 & 15.36 & 16.30 & 30.46 & 40.40 & 22.18 & 23.27 & 25.12 & 29.34 \\
& Ours & \textbf{47.83} & \textbf{51.09} & \underline{16.30} & \underline{16.30} & \textbf{25.39} & \textbf{26.33} & \textbf{63.58} & \textbf{68.87} & \textbf{65.09} & \textbf{66.18} & \textbf{64.55} & \textbf{67.14} \\
& Ours w/ RT & \underline{31.52} & \underline{35.87} & \textbf{16.74} & \textbf{16.74} & \underline{21.00} & \underline{22.26} & \underline{55.63} & \underline{58.94} & \underline{53.45} & \underline{53.82} & \underline{54.23} & \underline{55.63} \\
\bottomrule
\end{tabular}
\vspace{-3mm}
\end{table*}
\begin{table*}[t]
\centering
\caption{Triplet evaluation on the SceneFun3D~\cite{delitzas2024scenefun3d} and FunGraph3D~\cite{zhang2025open} datasets. Ours w/ RT denotes the real-time optimized variant. Best and second-best distinct results within each pose setting are highlighted in bold and underlined, respectively.}
\label{tab:triplet_evaluation_rt}
\scriptsize
\setlength{\tabcolsep}{3.0pt}
\begin{tabular}{llccccccccccccc}
\toprule
\multirow{3}{*}{Pose setting} & \multirow{3}{*}{Method}
& \multicolumn{6}{c}{SceneFun3D}
& \multicolumn{6}{c}{FunGraph3D}
& \multirow{3}{*}{FPS} \\
\cmidrule(lr){3-8} \cmidrule(lr){9-14}
& & \multicolumn{2}{c}{Node Assoc.}
& \multicolumn{2}{c}{Edge Pred.}
& \multicolumn{2}{c}{Overall Triplets}
& \multicolumn{2}{c}{Node Assoc.}
& \multicolumn{2}{c}{Edge Pred.}
& \multicolumn{2}{c}{Overall Triplets}
& \\
& & R@5 & R@10 & R@5 & R@10 & R@5 & R@10
& R@5 & R@10 & R@5 & R@10 & R@5 & R@10
& \\
\midrule
\multirow{3}{*}{GT pose}
& OpenFunGraph~\cite{zhang2025open} & \underline{27.19} & \underline{30.70} & \underline{53.23} & \textbf{91.43} & 14.47 & \textbf{28.07} & \underline{36.99} & \underline{41.10} & 48.15 & \underline{93.33} & \underline{17.81} & \underline{38.36} & 0.02 \\
& FunGraph~\cite{rotondi2025fungraph} & 18.42 & 19.74 & \textbf{85.71} & \underline{86.67} & \textbf{15.79} & \underline{17.11} & \textbf{41.78} & \textbf{42.12} & \textbf{96.72} & \textbf{96.75} & \textbf{40.41} & \textbf{40.75} & 0.06 \\
& KeySG~\cite{werby2025keysg} & \textbf{35.09} & \textbf{35.53} & 43.75 & 43.21 & \underline{15.35} & 15.35 & 19.52 & 23.29 & \underline{66.67} & 61.76 & 13.01 & 14.38 & 0.05 \\
\midrule
\multirow{5}{*}{Our pose}
& OpenFunGraph~\cite{zhang2025open} & 7.02 & 8.33 & 50.00 & \textbf{89.47} & 3.51 & 7.46 & 23.97 & 28.08 & 48.57 & 91.46 & 11.64 & 25.68 & -- \\
& FunGraph~\cite{rotondi2025fungraph} & \underline{9.65} & 10.09 & 72.73 & 73.91 & 7.02 & 7.46 & 27.74 & 28.42 & 97.53 & 97.59 & 27.05 & 27.74 & -- \\
& KeySG~\cite{werby2025keysg} & 4.39 & 4.39 & 30.00 & 30.00 & 1.32 & 1.32 & 4.79 & 8.22 & 92.86 & 87.50 & 4.45 & 7.19 & -- \\
& Ours & \textbf{11.40} & \textbf{11.84} & \underline{84.62} & 85.19 & \textbf{9.65} & \textbf{10.09} & \textbf{42.12} & \textbf{44.86} & \underline{97.56} & \underline{97.71} & \textbf{41.10} & \textbf{43.84} & \underline{0.38} \\
& Ours w/ RT & \underline{9.65} & \underline{10.53} & \textbf{86.36} & \underline{87.50} & \underline{8.33} & \underline{9.21} & \underline{35.62} & \underline{38.01} & \textbf{99.04} & \textbf{100.00} & \underline{35.27} & \underline{38.01} & \textbf{0.92} \\
\bottomrule
\end{tabular}
\end{table*}

On the FunGraph3D dataset, Ours w/ RT achieves an average speed of 1.027 FPS, yielding a 2.45$\times$ improvement over the unoptimized version. 
Table~\ref{tab:rt_optimization_runtime} compares the average per-frame runtime of each module on the 6kitchen/video0 sequence before and after runtime optimization. Ours w/ RT improves the overall speed from 0.377 FPS to 0.920 FPS, with the main acceleration coming from semantic information acquisition and online functional graph maintenance. First, the runtime of semantic information acquisition is reduced from 2040.06 ms/frame to 514.23 ms/frame, because RAM++, DeepSeek, and SAM3 are no longer repeatedly executed on non-keyframes, and the input resolution of SAM3 is reduced from 1008 to 672. Second, since the Online Functional Graph is updated only on keyframes, non-keyframes skip node association and edge association. As a result, the runtime of node association decreases from 117.66 ms/frame to 22.30 ms/frame, and that of edge association decreases from 12.76 ms/frame to 4.76 ms/frame. These results show that keyframe-level functional perception and functional graph construction can effectively reduce the computational burden introduced by per-frame online maintenance.

The accuracy loss caused by runtime optimization is overall controllable. On FunGraph3D, the ATE RMSE increases from 16.495 mm to 17.579 mm, with only a 1.084 mm increase. In terms of functional graph metrics, as shown in Table~~\ref{tab:node_evaluation_rt} and Table~\ref{tab:triplet_evaluation_rt}, Ours w/ RT still preserves the advantage of online functional graph maintenance over offline functional graph methods. Although Ours w/ RT reduces functional observations from non-keyframes compared with the full version, leading to some decreases in node and triplet metrics, its overall node recall and overall triplet recall remain clearly higher than those of offline functional graph methods such as OpenFunGraph, FunGraph, and KeySG under the same Ours-pose setting. More importantly, on FunGraph3D, Ours w/ RT still outperforms most offline results even when they use GT poses, indicating that keyframe-level functional perception does not destroy the core advantages brought by online node maintenance and temporal relation modeling. This advantage is achieved with substantially higher runtime efficiency. As shown in Table~\ref{tab:triplet_evaluation_rt}, Ours w/ RT reaches 0.92 FPS in this evaluation, whereas OpenFunGraph, FunGraph, and KeySG achieve only 0.02, 0.06, and 0.05 FPS, respectively.

Overall, the runtime optimization demonstrates that Functional-SLAM does not have to rely on dense semantic reasoning on every frame. By using keyframe-level functional perception, keyframe-level functional graph updates, and lower-resolution SAM3 inference, the system can substantially improve runtime efficiency while maintaining acceptable localization accuracy and functional graph quality. This provides a more practical efficiency-accuracy trade-off for future deployment.

\begin{figure}[t]
  \centering
   \includegraphics[width=1.0\linewidth]{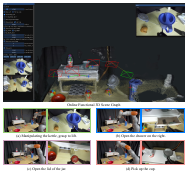}
   \caption{Real-robot manipulation examples using the online functional 3D scene graph. The top panel shows the online functional 3D scene graph constructed from the gripper-mounted ZED 2i camera, where functional nodes and edges are overlaid on the online 3D map. The bottom panels show multi-step natural-language manipulation tasks executed by a KUKA LBR iiwa 14 R820 robot with a Robotiq 2F-140 gripper: (a) manipulating the kettle and grasping it to lift, (b) opening the drawer on the right, (c) opening the lid of the jar, and (d) picking up the cup. 
   }
   \label{fig:robot} 
\end{figure}

\section{Manipulation Task Examples}

To verify that the functional 3D scene graph constructed by Functional-SLAM can support downstream robotic manipulation tasks, we conduct manipulation experiments on a real robotic platform. The platform consists of a KUKA LBR iiwa 14 R820 robot arm, a Robotiq 2F-140 gripper, and a ZED 2i RGB-D camera mounted at the gripper end-effector. During the experiments, the system uses only RGB images captured by the wrist-mounted camera for online perception and functional graph construction. Meanwhile, a calibration board is used to recover the scene scale, so that the 3D poses of functional nodes and interaction elements can be transformed into executable spatial coordinates for the robot arm.

At the task execution level, we build a large-language-model-based scene graph query interface, which converts natural-language manipulation instructions into structured queries over the functional 3D scene graph. Given an instruction, the system first parses the target object and the possible functional relations involved in the task. It then retrieves object nodes, interaction-element nodes, and functional links that satisfy the constraints from the currently constructed functional graph. If the query succeeds, the interface returns the 3D pose of the corresponding target object or interaction unit, which is then used by the robot execution module to perform actions such as grasping, pulling, opening, or closing.

The overall execution follows an online ``query--explore--execute'' strategy. After receiving a natural-language instruction, the system first checks whether the current functional 3D scene graph already contains the functional structure required for the task, including the target object, the target interaction element, and their functional relation chain. If such a structure exists, the robot directly executes the task using the queried 3D pose. Otherwise, the system does not wait for a complete offline reconstruction. Instead, it continues actively observing the environment with the wrist-mounted camera and incrementally updates the functional scene graph during exploration. Once nodes and functional relations satisfying the task constraints appear in the graph, the system immediately triggers task execution.

In the experiment, we simulate continuous task execution for a robot operating in a realistic home environment, as shown in Fig.~\ref{fig:robot}. First, the user issues the instruction, ``Manipulate the kettle: grasp it to lift.'' At this moment, the current functional graph does not yet contain the kettle and its graspable interaction element required by the instruction. The robot therefore continues to explore and construct the functional scene graph. Once the system online constructs the kettle, its handle, and the functional relation ``grasp to lift or pour,'' the query interface returns the 3D pose of the corresponding handle, allowing the robot to grasp and lift it. The user then issues subsequent instructions, such as ``Open the drawer on the right,'' ``Open the lid of the jar,'' and ``Pick up the cup.'' Since the previous exploration has already maintained related functional structures such as drawer--knob, jar--lid, and cup--handle in the functional graph, the system can directly retrieve the target nodes and functional relations from the current graph and pass the corresponding 3D poses to the robot execution module for manipulation.

This experiment shows that Functional-SLAM can satisfy the requirements of immediate interaction. Unlike functional scene graph methods that rely on known poses or offline reconstruction followed by post-processing, Functional-SLAM continuously maintains functional nodes, interaction elements, and their functional relations during exploration. As a result, the robot can immediately query the current graph state when a task arrives. When the graph lacks the necessary functional structure, the system can continue exploration and incrementally complete the graph until the conditions for task execution are satisfied.

The complete multi-step manipulation process is provided in the supplementary video.

\begin{figure}[t]
  \centering
   \includegraphics[width=1.0\linewidth]{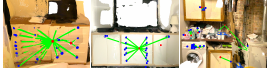}
   \caption{Representative failure cases of Functional-SLAM. The examples are taken from SceneFun3D sequences with noticeable camera-pose drift and unstable local reconstruction. 
   }
   \label{fig:FailureCases} 
\end{figure}

\section{Failure Cases and Limitations}

Although Functional-SLAM significantly improves the stability of functional graph construction under online pose estimation, it still has failure cases in extreme scenarios.

The first type of failure mainly comes from large camera pose drift. Anchor-keyframe-based geometry and functional-context-constrained node association can partially alleviate geometric misalignment caused by online pose drift. They keep the geometry of functional nodes synchronized with back-end pose optimization and use historical functional ownership to constrain the cross-frame identities of small interaction elements. However, when pose drift becomes severe, the 3D support back-projected from 2D masks can be noticeably shifted. This issue is especially critical for small interaction elements such as knobs, buttons, and handles, which have small spatial extents, sparse point clouds, and high sensitivity to pose errors. In this case, the 3D centers, projected bounding boxes, and local geometric support of candidate nodes may become unstable, leading to failed or incorrect cross-frame associations. This problem is more evident on the SceneFun3D dataset, where most sequences suffer from large camera pose drift and insufficient observations. Fig.~\ref{fig:FailureCases} shows several representative results. When the camera trajectory has obvious drift, local reconstruction is unstable, and observations of small interaction elements are sparse, the system may still produce node association failures, duplicate nodes, or incorrect functional edge connections.

The second issue is related to the sensitivity of the evaluation protocol itself. Functional graph triplet recall imposes strict geometric matching requirements on the two endpoint nodes. A triplet is counted as correct only when the interaction-element endpoint, the interacted-object endpoint, and the functional relation are all successfully matched. Therefore, even if the system predicts a semantically correct functional relation, the entire triplet is still regarded as incorrect if the 3D bounding box of a small interaction element is too small, its point cloud is too sparse, or pose errors prevent sufficient overlap with the ground-truth node bounding box. This phenomenon is particularly evident on SceneFun3D\_Graph. In our experiments, we observe that our method usually achieves high Edge Prediction recall on triplets whose endpoints have already been associated, indicating that the semantics of functional relations can be predicted relatively stably. However, the Node Association recall is lower, suggesting that the main bottleneck lies in the 3D association failure of small interaction elements and object endpoints, rather than in incorrect relation-text prediction. This also explains why, in sequences with severe pose drift, the ability to predict functional relations does not necessarily translate into improved final triplet recall.

The third limitation comes from scene state changes. The current Functional-SLAM continuously maintains functional nodes and their relations as online map states, but it does not explicitly model object state changes caused by robot manipulation. For example, a door may be opened or closed, a drawer may be pulled out, a switch may change its state, and a movable object may be relocated. These changes affect not only local geometry but also the functional relations between interaction elements and objects. The current system mainly targets static or quasi-static exploration and cannot yet handle such dynamic graph updates induced by manipulation actions. Future work should introduce a state-aware functional graph representation, enabling the system to model and update the state changes of manipulated objects and the corresponding changes in functional relations.

\section{Supplementary Evaluation with a Unified SAM3 Front End}

The functional graph recall results in the main paper already use the same LLM as ours to minimize the influence of different relation reasoning models. To further verify that the performance improvement of Functional-SLAM does not mainly come from differences in the detection or segmentation front end, we additionally replace the detection/segmentation modules of the baseline methods with a SAM3 setting consistent with ours as much as possible in the supplementary material, while keeping their subsequent graph construction and relation generation pipelines unchanged.

For different baseline methods, we adopt replacement strategies according to their original visual perception front ends. OpenFunGraph and KeySG originally use GroundingDINO~\cite{liu2024grounding}+SAM~\cite{kirillov2023segment} as their open-vocabulary detection and segmentation front end. Therefore, we replace their detection and segmentation front ends with SAM3 outputs, while keeping their subsequent graph generation and relation reasoning pipelines unchanged. In the final system of FunGraph, common objects are detected by YOLO-Worldv8.2~\cite{cheng2024yolo}, while functional interaction elements are detected by RT-DETR~\cite{zhao2024detrs} fine-tuned on 2D data generated from SceneFun3D projections, and the detected boxes are then sent to SAM2~\cite{ravi2025sam} to generate masks. Therefore, for FunGraph, we only replace the box-prompted segmenter from SAM2 to SAM3~\cite{carion2025sam3segmentconcepts}, while keeping the detector unchanged, to avoid altering its original detection design.

Tables~\ref{tab:node_evaluation_sam3} and~\ref{tab:triplet_evaluation_sam3} report the node recall and triplet recall results after unifying the SAM3 front end, respectively. Under the same Ours-pose setting, Functional-SLAM still achieves the best overall node recall and overall triplet recall on both SceneFun3D and FunGraph3D. This indicates that our advantage does not mainly come from using a stronger detector or segmenter, but from the online functional graph maintenance mechanism itself. Specifically, the anchor-keyframe-based geometry keeps functional node geometry synchronized with back-end pose optimization, the functional-context-constrained node association reduces cross-instance confusion among nodes, and the temporal relation posterior suppresses single-frame relation ambiguity through multi-frame evidence. Therefore, even after unifying the SAM3 front end, Functional-SLAM can still construct more accurate and more stable online functional graphs.

It is worth noting that replacing the original front ends of the baseline methods with SAM3 does not necessarily improve recall. This is because the downstream graph construction pipelines of offline functional graph methods are usually highly coupled with the output distributions of their original front ends, and do not depend only on mask quality. For example, these pipelines also rely on the number of candidate proposals, box/mask scales, parent-child containment relations, spatial adjacency, size filtering, node merging or splitting strategies, label generation, and relation reasoning thresholds. The output distribution of SAM3 is not fully consistent with that of GroundingDINO+SAM or SAM2. Without re-tuning these downstream rules, replacing the front end may instead degrade graph construction quality. For OpenFunGraph and KeySG, replacing GroundingDINO+SAM changes not only the mask generation process, but also the mechanism for producing open-vocabulary proposals. GroundingDINO generates category-related candidates according to text prompts, whereas the candidate distribution of SAM3 may not be fully aligned with these text prompts or the benchmark annotation vocabulary. As a result, some objects or small functional parts that could originally be proposed by GroundingDINO may not be recalled as stably under the SAM3 front end, thereby affecting node and relation evaluation. For FunGraph, although the detection boxes remain unchanged and only SAM2 is replaced with SAM3, changes in the mask distribution can still affect subsequent containment reasoning, parent-child assignment, and functional edge generation.

This effect becomes more evident under the Ours-pose setting. Compared with GT poses, online estimated poses introduce 2D-to-3D alignment errors. SAM3 often produces tighter masks, which may benefit segmentation boundaries under clean poses. However, in the presence of pose errors, tighter masks can further reduce the valid 3D points of small interaction elements, leading to unstable 3D bounding boxes, failed parent-object association, or failed triplet endpoint matching. Therefore, the unified SAM3 front-end experiment does not imply that SAM3 universally improves all offline functional graph methods. Instead, it shows that even when the detection and segmentation front ends are unified as much as possible, Functional-SLAM still maintains a clear advantage, and its performance gain mainly comes from online geometric synchronization, persistent node maintenance, and temporal stabilization of functional relations.

\begin{table*}[t]
\centering
\caption{Node evaluation on the SceneFun3D~\cite{delitzas2024scenefun3d} and FunGraph3D~\cite{zhang2025open} datasets. Baseline methods use SAM3 as the detection and segmentation module. Best and second-best distinct results within each pose setting are highlighted in bold and underlined, respectively.}
\label{tab:node_evaluation_sam3}
\scriptsize
\setlength{\tabcolsep}{3.2pt}
\begin{tabular}{llcccccccccccc}
\toprule
\multirow{3}{*}{Pose setting} & \multirow{3}{*}{Method}
& \multicolumn{6}{c}{SceneFun3D}
& \multicolumn{6}{c}{FunGraph3D} \\
\cmidrule(lr){3-8} \cmidrule(lr){9-14}
& & \multicolumn{2}{c}{Objects}
& \multicolumn{2}{c}{Inter. Elements}
& \multicolumn{2}{c}{Overall Nodes}
& \multicolumn{2}{c}{Objects}
& \multicolumn{2}{c}{Inter. Elements}
& \multicolumn{2}{c}{Overall Nodes} \\
& & R@3 & R@10 & R@3 & R@10 & R@3 & R@10
& R@3 & R@10 & R@3 & R@10 & R@3 & R@10 \\
\midrule
\multirow{3}{*}{GT pose}
& OpenFunGraph~\cite{zhang2025open}
& \underline{33.70} & \underline{38.04}
& 47.58 & \underline{53.30}
& 43.57 & \underline{48.90}
& \underline{37.75} & \underline{43.05}
& 40.73 & 46.91
& 39.67 & \underline{45.54} \\
& FunGraph~\cite{rotondi2025fungraph}
& \textbf{35.87} & \textbf{40.22}
& \textbf{64.32} & \textbf{66.96}
& \textbf{56.11} & \textbf{59.25}
& \textbf{49.01} & \textbf{54.30}
& \textbf{54.91} & \textbf{56.00}
& \textbf{52.82} & \textbf{55.40} \\
& KeySG~\cite{werby2025keysg}
& 25.00 & 32.61
& \underline{52.86} & 52.86
& \underline{44.83} & 47.02
& 31.13 & 35.76
& \underline{46.91} & \underline{47.64}
& \underline{41.31} & 43.43 \\
\midrule
\multirow{4}{*}{Ours pose}
& OpenFunGraph~\cite{zhang2025open}
& 22.83 & 25.00
& 13.22 & 15.86
& 15.99 & 18.50
& 34.44 & 39.07
& 25.45 & 31.27
& 28.64 & 34.04 \\
& FunGraph~\cite{rotondi2025fungraph}
& \underline{28.26} & \underline{30.43}
& \underline{15.86} & \textbf{16.74}
& \underline{19.44} & \underline{20.69}
& \underline{47.02} & \underline{50.99}
& \underline{35.27} & \underline{37.09}
& \underline{39.44} & \underline{42.02} \\
& KeySG~\cite{werby2025keysg}
& 17.39 & 23.91
& 7.49 & 8.37
& 10.34 & 12.85
& 24.50 & 30.46
& 15.27 & 15.27
& 18.54 & 20.66 \\
& Ours
& \textbf{47.83} & \textbf{51.09}
& \textbf{16.30} & \underline{16.30}
& \textbf{25.39} & \textbf{26.33}
& \textbf{63.58} & \textbf{68.87}
& \textbf{65.09} & \textbf{66.18}
& \textbf{64.55} & \textbf{67.14} \\
\bottomrule
\end{tabular}
\vspace{-3mm}
\end{table*}
\begin{table*}[t]
\centering
\caption{Triplet evaluation on the SceneFun3D~\cite{delitzas2024scenefun3d} and FunGraph3D~\cite{zhang2025open} datasets. Baseline methods use SAM3 as the detection and segmentation module. Best and second-best distinct results within each pose setting are highlighted in bold and underlined, respectively.}
\label{tab:triplet_evaluation_sam3}
\scriptsize
\setlength{\tabcolsep}{3.2pt}
\begin{tabular}{llcccccccccccc}
\toprule
\multirow{3}{*}{Pose setting} & \multirow{3}{*}{Method}
& \multicolumn{6}{c}{SceneFun3D}
& \multicolumn{6}{c}{FunGraph3D} \\
\cmidrule(lr){3-8} \cmidrule(lr){9-14}
& & \multicolumn{2}{c}{Node Assoc.}
& \multicolumn{2}{c}{Edge Pred.}
& \multicolumn{2}{c}{Overall Triplets}
& \multicolumn{2}{c}{Node Assoc.}
& \multicolumn{2}{c}{Edge Pred.}
& \multicolumn{2}{c}{Overall Triplets} \\
& & R@5 & R@10 & R@5 & R@10 & R@5 & R@10
& R@5 & R@10 & R@5 & R@10 & R@5 & R@10 \\
\midrule
\multirow{3}{*}{GT pose}
& OpenFunGraph~\cite{zhang2025open}
& \textbf{28.07} & \textbf{29.82}
& 67.19 & 82.35
& \textbf{18.86} & \textbf{24.56}
& \underline{38.01} & \underline{39.73}
& 67.57 & 81.90
& \underline{25.68} & \underline{32.53} \\
& FunGraph~\cite{rotondi2025fungraph}
& \underline{16.67} & \underline{17.98}
& \underline{97.37} & \underline{97.56}
& \underline{16.23} & \underline{17.54}
& \textbf{42.47} & \textbf{43.84}
& \underline{96.77} & \textbf{96.88}
& \textbf{41.10} & \textbf{42.47} \\
& KeySG~\cite{werby2025keysg}
& 13.60 & 16.23
& \textbf{100.00} & \textbf{100.00}
& 13.60 & 16.23
& 12.67 & 18.49
& \textbf{97.30} & \underline{94.44}
& 12.33 & 17.47 \\
\midrule
\multirow{4}{*}{Ours pose}
& OpenFunGraph~\cite{zhang2025open}
& 7.02 & 7.89
& 68.75 & 83.33
& 4.82 & 6.58
& 24.66 & 27.05
& 68.06 & 82.28
& 16.78 & 22.26 \\
& FunGraph~\cite{rotondi2025fungraph}
& \underline{9.65} & \underline{10.09}
& 72.73 & 73.91
& \underline{7.02} & \underline{7.46}
& \underline{28.08} & \underline{28.42}
& \underline{97.56} & 97.59
& \underline{27.40} & \underline{27.74} \\
& KeySG~\cite{werby2025keysg}
& 1.75 & 1.75
& \textbf{100.00} & \textbf{100.00}
& 1.75 & 1.75
& 3.08 & 4.79
& \textbf{100.00} & \textbf{100.00}
& 3.08 & 4.79 \\
& Ours
& \textbf{11.40} & \textbf{11.84}
& \underline{84.62} & \underline{85.19}
& \textbf{9.65} & \textbf{10.09}
& \textbf{42.12} & \textbf{44.86}
& \underline{97.56} & \underline{97.71}
& \textbf{41.10} & \textbf{43.84} \\
\bottomrule
\end{tabular}
\vspace{-3mm}
\end{table*}

\section{Implementation Details}

\subsection{Functional Graph Representation}

To avoid ambiguity, we further clarify the functional graph representation used by Functional-SLAM in the supplementary material. In the main paper, following OpenFunGraph, we represent the online functional graph as $G=(U,O,R)$, where $O$ denotes manipulable objects, $U$ denotes operable interaction elements, and $R$ denotes functional relations between objects and interaction elements. This representation summarizes the core information that the final functional graph should capture: which objects and interaction elements exist in the scene, and how the interaction elements act on the corresponding objects. In the implementation, we further adopt the $O$-$C$-$U$ hierarchical structure inspired by HHOpenFunGraph as the internal maintenance form. Here, $O$ denotes manipulable objects, $C$ denotes functional carriers, and $U$ denotes directly operable interaction units. A functional carrier $C$ is an intermediate functional part between an object and an interaction unit, describing the local structure to which the interaction unit is attached or on which it acts. For example, in cabinet-drawer-handle, cabinet is the object $O$, drawer is the functional carrier $C$, and handle is the interaction unit $U$. In appliance-panel-button, appliance is the object $O$, panel is the functional carrier $C$, and button is the interaction unit $U$.

The main purpose of using the $O$-$C$-$U$ hierarchy in the implementation is to provide richer and more discriminative structural cues for functional-topology-assisted loop closure. Compared with directly maintaining only $O$-$U$ connections between objects and interaction units, the $O$-$C$-$U$ hierarchy explicitly preserves the functional carrier $C$. As a result, each local functional structure contains not only ``which interaction unit acts on which object'', but also ``through which intermediate functional part the interaction unit acts on the object''. Therefore, a local region can form a finer-grained functional topological signature from the object, functional carrier, interaction unit, and their hierarchical connections. This hierarchical topology is particularly useful for loop candidate generation. In scenes with repeated appearances, weak textures, or similar geometric structures, relying only on visual features or simple $O$-$U$ relations can easily introduce ambiguity. In contrast, the $O$-$C$-$U$ structure can distinguish finer functional patterns, such as ``cabinet-drawer-handle'' versus ``cabinet-door-knob'', or multiple handles corresponding to different drawers under the same cabinet. In other words, even if multiple interaction units have the same label and share the same object node, they may still form different local topological contexts through different functional carriers $C$, thereby providing more discriminative functional structural constraints for loop closure.

It should be noted that the $O$-$C$-$U$ hierarchy does not change the functional graph definition $G=(U,O,R)$ used in the main paper. It is only a refined maintenance form for functional relations between objects and interaction elements. Once the system establishes an $O$-$C$-$U$ hierarchical relation, a functional connection is naturally formed between the object $O$ and the interaction unit $U$. For interaction units without an explicit intermediate functional carrier, the system directly establishes an $O$-$U$ connection. Therefore, during final readout or when aligning with the OpenFunGraph evaluation protocol, the $O$-$C$-$U$ chain can be collapsed into a functional relation between an object and an interaction element, while the $C$ layer is preserved during internal maintenance to enhance node association, relation stabilization, and loop closure.

\subsection{Functional-Context-Constrained Node Association}

Node association in the online functional graph aims to stably map functional nodes observed in the current frame to existing persistent nodes. Different from standard object-level association, interaction elements are usually small, have weak geometric support, and may have multiple visually and semantically similar candidates within the same local region, such as multiple handles or knobs on adjacent drawers, or caps on densely arranged bottles. Therefore, relying only on geometric projection or open-vocabulary labels can easily cause cross-instance confusion. To address this issue, Functional-SLAM explicitly introduces functional context constraints into node association. In this way, a current observation is required to be consistent not only with the candidate node itself, but also with the historical functional parent of that node and the local relation evidence in the current frame.

Given a current-frame node observation $n_i^t$ and the persistent node set $V_{t-1}$ from the previous time step, the system first performs role-wise matching. That is, objects $O$, functional carriers $C$, and interaction units $U$ are associated only with persistent nodes of the same role. For a current observation $n_i^t$ and a same-role candidate node $v_j$, the system recovers the predicted geometry of the candidate node under the latest SLAM state, including its world-coordinate center and its 2D bounding box projected onto the current frame. It then constructs the candidate matching score as
\begin{equation}
\small
S_{ij}^t =
\phi_{\mathrm{geo}}(n_i^t,v_j)
+
\phi_{\mathrm{sem}}(n_i^t,v_j)
+
\phi_{\mathrm{ctx}}(n_i^t,v_j,B_t).
\end{equation}
Here, $\phi_{\mathrm{geo}}$ measures consistency in terms of projection overlap, 3D center distance, and scale-adaptive gating based on the predicted node geometry under the latest optimized pose. $\phi_{\mathrm{sem}}$ measures the consistency between the open-vocabulary label and the historical semantic statistics of the candidate node. $\phi_{\mathrm{ctx}}$ uses the current relation evidence $B_t$ to constrain the functional ownership of the observation.

In the default implementation, the score is computed as
\begin{equation}
\small
S_{ij}^{t}
=
0.45 O_{ij}
+
0.15 D_{ij}
+
0.30 L_{ij}
+
0.10 T_{ij}
+
0.10 C_{ij}.
\end{equation}
Here, $O_{ij}$ and $D_{ij}$ form the geometric consistency term $\phi_{\mathrm{geo}}$, $L_{ij}$ and $T_{ij}$ form the semantic consistency term $\phi_{\mathrm{sem}}$, and $C_{ij}$ forms the functional context consistency term $\phi_{\mathrm{ctx}}$. The weights follow a conservative principle: projection overlap $O_{ij}$ and label consistency $L_{ij}$ are the most direct cues for node identity, and are therefore assigned larger weights; the 3D distance $D_{ij}$ is affected by online pose and depth noise, and is thus assigned a smaller weight; the semantic subtype $T_{ij}$ and functional context $C_{ij}$ mainly help distinguish adjacent interaction elements of the same category, and are therefore used as auxiliary terms in the score.

The geometric overlap term $O_{ij}$ measures the consistency between the projected bounding box of the candidate node and the bounding box of the current observation. Based on the anchor-synchronized geometry of the candidate node, the system recovers the node geometry under the current SLAM state and projects it onto the current frame to compute the 2D IoU.

The distance consistency term $D_{ij}$ measures the distance between the 3D center of the current observation and the predicted center of the candidate node. Let this distance be $d_{ij}$. Then,
\begin{equation}
\small
D_{ij}=\max(0,1-d_{ij}/\tau_j).
\end{equation}
The distance threshold $\tau_j$ is adaptively set according to the scale and role of the candidate node. Specifically, the system first uses $1.2$ times the diagonal length of the 3D bounding box of the candidate node as the initial threshold, and then clamps it within the lower and upper bounds corresponding to that role:
\begin{equation}
\small
\tau_j =
\min\left(
\tau_{\max}(r_j),
\max\left(
1.2\operatorname{diag}(v_j),
\tau_{\min}(r_j)
\right)
\right),
\end{equation}
where $r_j\in\{O,C,U\}$ denotes the role of the candidate node $v_j$. By default, the threshold ranges for objects $O$, functional carriers $C$, and interaction units $U$ are $[0.15,1.20]$ m, $[0.08,0.60]$ m, and $[0.04,0.35]$ m, respectively. This setting is used because objects are usually larger and can tolerate larger projection and depth errors, whereas interaction units such as handles, knobs, and buttons are small and often densely distributed across neighboring instances, requiring stricter distance constraints.

The semantic label term $L_{ij}$ measures the consistency between the current observation label and the historical labels of the candidate node. The current observation label comes from the open-vocabulary detection result in the current frame. The candidate node maintains historical label counts and derives a stable dominant label from them. If the current observation label is the same as the dominant label of the candidate node, $L_{ij}=1.0$. If it has appeared in the historical label set of the node, $L_{ij}=0.7$. Otherwise, $L_{ij}=0.1$. This design tolerates slight label fluctuations in open-vocabulary detection while still prioritizing the long-term stable node label.

The semantic subtype term $T_{ij}$ is used to distinguish nodes that have the same label but different functional ownership. For example, a handle may belong to a drawer or a door. The system forms subtypes such as ``handle@drawer'' or ``handle@door'' according to the local functional relations in the current frame. If the current observation subtype is consistent with the historical subtype of the candidate node, $T_{ij}=1.0$. If they conflict with each other, $T_{ij}=0$. If either the current observation or the candidate node lacks a reliable subtype, $T_{ij}=0.5$. Since the subtype depends on local relation reasoning, the system assigns it only an auxiliary weight of $0.10$, preventing it from overriding geometric and dominant-label evidence when it is unstable.

The functional context term is mainly used for nodes with parent nodes. Object nodes $O$ do not have parent nodes, so $C_{ij}$ and $P_{ij}$ are not enabled for $O$. The parent of a functional carrier node $C$ is usually an object $O$, while the parent of an interaction unit $U$ is usually a functional carrier $C$, and may also be directly an object $O$. In the current implementation, the positive parent consistency term $C_{ij}$ is mainly used for interaction units $U$, because interaction units with the same name are most likely to suffer from identity confusion among adjacent doors, drawers, or devices. For functional carriers $C$, the parent-object relation is mainly stabilized by the subsequent relation posterior, while its historical parent node can still participate in the association constraint through the penalty term $P_{ij}$. For interaction units, the system compares whether the parent evidence observed in the current frame is consistent with the historical parent node of the candidate. The parent consistency score is defined as
\begin{equation}
\small
C_{ij}
=
0.80G_{ij}^{p}
+
0.15R_{ij}^{p}
+
0.05L_{ij}^{p},
\end{equation}
where $G_{ij}^{p}$ denotes parent-node geometric consistency, which combines parent projection overlap and 3D distance; $R_{ij}^{p}$ denotes whether the parent roles are consistent; and $L_{ij}^{p}$ denotes whether the parent labels are consistent. A larger weight is assigned to parent geometry because, when multiple parent nodes of the same category coexist, the spatial location of the parent is more discriminative than its textual label. If the current observation has no reliable parent evidence, or the candidate node has not yet formed a historical parent, then $C_{ij}=0$, and no additional score is added.

Finally, the system constructs a valid candidate score matrix within each role and solves the one-to-one optimal matching using the Hungarian algorithm. Matched observations are used to update the corresponding persistent nodes. Unmatched observations are initialized as new candidate nodes and continue to receive geometric and functional context verification in subsequent frames. With this design, functional context directly participates in node identity association, rather than being used only as a post-processing step for relation correction, thereby reducing cross-instance confusion among adjacent interaction elements of the same category.

\subsection{Functional-Topology-Assisted Loop Closure}

Functional-SLAM introduces the online functional graph as a structured map memory on top of the visual retrieval module of MASt3R-SLAM~\cite{murai2025mast3r}, so as to supplement loop candidates that may be missed by visual retrieval. For a current query keyframe $k$ and a historical keyframe $k'$, the functional topology module only computes candidate similarity and provides additional candidates. The final candidates still need to pass geometric verification, and only verified candidates are added to the back-end optimization as loop constraints.

For each keyframe $k$, the system constructs a local functional topological signature $\Gamma_k$ from the online functional graph. Specifically, for each visible stable node $v_i$ in the keyframe, the system selects its two nearest stable neighbors $v_{i1}$ and $v_{i2}$ in 3D space, forming a three-node graphlet:
\begin{equation}
\small
g_i=\{v_i,v_{i1},v_{i2}\}.
\end{equation}
The functional topological signature of keyframe $k$ is then defined as the set of all graphlets:
\begin{equation}
\small
\Gamma_k=\{g_i\}.
\end{equation}
Each graphlet stores the persistent node IDs, roles, open-vocabulary labels, and 3D centers of its nodes. The node roles include object $O$, functional carrier $C$, and interaction unit $U$. An interaction unit is used for topology matching only when it participates in a maintained functional chain or a direct functional relation, so as to avoid interference from isolated low-confidence interaction elements. In the implementation, each keyframe keeps at most 64 graphlets, and at most the first 24 graphlets are used when comparing two frames, in order to control the computational cost.

Given $g\in\Gamma_k$ and $h\in\Gamma_{k'}$, the system first computes their semantic compatibility $S_{\mathrm{sem}}(g,h)$. Let $g=\{v_1,v_2,v_3\}$ and $h=\{u_1,u_2,u_3\}$. Since there is no fixed node order inside a three-node graphlet, the system enumerates all one-to-one correspondences among the three nodes and selects the correspondence with the highest semantic compatibility. Let $\Pi_3$ denote all permutations of three nodes. Then,
\begin{equation}
\small
S_{\mathrm{sem}}(g,h)
=
\max_{\sigma\in\Pi_3}
\frac{1}{3}
\sum_{a=1}^{3}
\chi(v_a,u_{\sigma(a)}).
\end{equation}
Here, $\chi(v,u)$ is the node compatibility function. If $v$ and $u$ share the same persistent node ID, then $\chi(v,u)=1$. If they are not the same persistent node but have the same role and the same open-vocabulary label, we also set $\chi(v,u)=1$. Otherwise, $\chi(v,u)=0$. Therefore, $S_{\mathrm{sem}}$ jointly uses persistent node identity across keyframes, functional roles, and open-vocabulary semantics. This definition assigns the highest compatibility when the same node is repeatedly observed, while still supporting matching when the two graphlets do not share node IDs but have consistent local functional semantics.

In addition to semantic compatibility, the system also computes the shape consistency of graphlets. Let the three edge lengths in graphlet $g$ be sorted as $\tilde d_1,\tilde d_2,\tilde d_3$. Its normalized shape vector is defined as
\begin{equation}
\small
\eta(g)
=
\frac{1}{\tilde d_1+\tilde d_2+\tilde d_3+\epsilon}
(\tilde d_1,\tilde d_2,\tilde d_3).
\end{equation}
The shape consistency between two graphlets is defined as
\begin{equation}
\small
S_{\mathrm{shape}}(g,h)
=
\frac{1}{1+\left|\eta(g)-\eta(h)\right|_1}.
\end{equation}
This term encodes only the relative distance ratios among local nodes and does not depend on the global coordinate direction. Therefore, it is suitable as a local topological shape constraint for loop candidate generation. The final topological matching score between two graphlets is defined as
\begin{equation}
\small
w(g,h)
=
S_{\mathrm{sem}}(g,h)S_{\mathrm{shape}}(g,h).
\end{equation}
This definition means that a high topological matching score is obtained only when two graphlets are consistent in both node semantics and local geometric shape.

For each graphlet $g$ in the current keyframe $k$, the system finds its best match in the graphlet set of the historical keyframe $k'$:
\begin{equation}
\small
h_g
=
\arg\max_{h\in\Gamma_{k'}} w(g,h).
\end{equation}
The object-level topological similarity between the two keyframes is then defined as
\begin{equation}
\small
S_G(k,k')
=
\frac{1}{|\Gamma_k|}
\sum_{g\in\Gamma_k}
w(g,h_g).
\end{equation}
Thus, $S_G$ measures whether each local functional topological structure in the current keyframe can find a semantically and geometrically similar local structure in the historical keyframe. It mainly reflects the consistency of the roles, labels, and relative spatial layout of stable nodes.

$S_G$ measures only whether local node topologies are similar, but does not explicitly check whether functional relations are preserved. Therefore, the system further computes the functional relation preservation ratio $S_F(k,k')$. For a graphlet $g$, let $E_g$ denote the set of functional edges induced by the maintained functional graph. In the $O$-$C$-$U$ hierarchy, a complete functional chain can be decomposed into directed edges $U\rightarrow C$ and $C\rightarrow O$; for cases without an intermediate functional carrier, the direct relation is represented as $U\rightarrow O$. Given the best node correspondence between $g$ and $h_g$, a functional edge in $g$ is considered preserved in $h_g$ if the corresponding endpoints, edge type, and direction are all preserved.

For a matched graphlet, the functional relation preservation ratio is defined as
\begin{equation}
\small
\psi(g,h_g)
=
\frac{
|\{e\in E_g \mid e \ \mathrm{is\ preserved\ in}\ h_g\}|
}{
|E_g|
}.
\end{equation}
When $g$ contains no functional edge, this graphlet does not provide functional relation constraints. For all valid graphlets, the system uses the topological matching quality $w(g,h_g)$ as the weight and computes the keyframe-level functional relation preservation ratio as
\begin{equation}
\small
S_F(k,k')
=
\frac{
\sum_{g\in\Gamma_k}
w(g,h_g)\psi(g,h_g)
}{
\sum_{g\in\Gamma_k}
w(g,h_g)
}.
\end{equation}
Therefore, $S_F$ does not independently compare the overlap between functional edge sets. Instead, under the established local graphlet correspondences, it checks whether the directions and hierarchical relations of functional edges are preserved. This avoids false loop retrieval caused by a small number of same-label edges, while rewarding historical keyframes that are consistent in both local topology and functional relations.

The final functional topology similarity is defined as
\begin{equation}
\small
S_{\mathrm{FT}}(k,k')
=
S_G(k,k')
\left(1+S_F(k,k')\right).
\end{equation}
Here, $S_F$ is a positive enhancement term for $S_G$, rather than an independent retrieval condition. In other words, a candidate keyframe must first have a certain degree of local topological similarity. On this basis, if functional edges are also preserved in the matched structures, its ranking is further improved. If $S_G(k,k')=0$, then $S_F(k,k')$ cannot independently produce a high-scoring candidate.

Candidate generation follows the decision form in Eq.~(6) of the main paper:
\begin{equation}
\small
S_G(k,k')>\tau_G
\quad \mathrm{or} \quad
S_{\mathrm{FT}}(k,k')>\tau_{\mathrm{FT}}.
\end{equation}
If the condition is satisfied, the historical keyframe $k'$ can be used as a supplementary candidate from functional topology. In the default implementation, we set $\tau_G=\tau_{\mathrm{FT}}=0$, and retain the top-$K$ candidates with the highest $S_G$ and $S_{\mathrm{FT}}$ scores, respectively, with $K=3$ by default. Meanwhile, to avoid treating temporally adjacent local keyframes as long-term loop candidates, the system requires $k'$ and the current keyframe $k$ to be separated by at least 10 keyframes. It should be emphasized that these functional topology candidates are used only to supplement candidates that may be missed by MASt3R-SLAM visual retrieval. They do not replace or remove the original visual candidates.

All functional topology candidates still need to pass geometric verification. After merging visual retrieval candidates and supplementary functional topology candidates, the system sends them to MASt3R geometric matching and factor-graph verification. Functional topology only determines which historical keyframes are worth attempting for geometric verification, but does not directly determine whether a loop closure is valid. A candidate becomes a true loop constraint only when geometric verification succeeds and a factor edge can be added. Otherwise, the candidate does not participate in back-end optimization. This design enables functional topology to improve loop candidate recall while preserving the geometric reliability of the final loop constraints.

\subsection{Remote Relations}

In addition to local functional relations between interaction elements and their associated objects or functional carriers, Functional-SLAM also maintains remote functional relations between spatially separated objects. Remote relations describe non-contact functional dependencies, such as a switch controlling a light or a socket powering an appliance. Different from local relations, remote relations usually cannot be reliably inferred from 2D overlap, mask containment, or 3D proximity alone. Therefore, the system first treats remote relations as label-level candidates, and then promotes them to instance-level relations through node association and multi-frame evidence accumulation.

For the current frame, the open-vocabulary reasoning module first produces a label-level remote relation candidate:
\begin{equation}
\small
e_t^{\mathrm{remote}} = (\ell_s, \rho, \ell_d),
\end{equation}
where $\ell_s$ and $\ell_d$ denote the source and target object labels, respectively, and $\rho$ denotes the type of remote relation. Let the corresponding detections be $n_s^t$ and $n_d^t$. The node association function $\pi_t(\cdot)$ maps them to persistent nodes, and the remote relation is then promoted to an instance-level endpoint pair:
\begin{equation}
\small
p = (v_s, v_d) = \left(\pi_t(n_s^t), \pi_t(n_d^t)\right).
\end{equation}

For an endpoint pair $p$, we compute its current-frame support score as
\begin{equation}
\small
\omega_t(p,\rho)
=
\lambda_{\mathrm{det}}
+
\lambda_{\mathrm{conf}}\min(s_s^t,s_d^t)
+
\lambda_{\mathrm{cov}} I_t(v_s,v_d)
+
\lambda_{\mathrm{mat}}\frac{m(v_s)+m(v_d)}{2}
+
\lambda_{\mathrm{dist}} D_\rho(d(v_s,v_d)).
\end{equation}
Here, $s_s^t$ and $s_d^t$ are the detection confidences of the source and target endpoints, $I_t(v_s,v_d)$ indicates whether the two endpoints are co-visible in the current frame, $m(\cdot)$ denotes the geometric maturity of an endpoint node, $d(v_s,v_d)$ is the 3D distance between the two endpoints, and $D_\rho(\cdot)$ is a relation-dependent distance prior. The default weights are set as
\begin{equation}
\small
\lambda_{\mathrm{det}}=1.0,\quad
\lambda_{\mathrm{conf}}=1.0,\quad
\lambda_{\mathrm{cov}}=0.35,\quad
\lambda_{\mathrm{mat}}=0.35,\quad
\lambda_{\mathrm{dist}}=0.45.
\end{equation}

These parameters measure the support strength of the current frame for a specific instance-level endpoint pair, rather than serving as hard thresholds for committing a remote edge from a single frame. The detection and confidence terms are assigned larger weights because remote relations first come from label-level candidates produced by open-vocabulary reasoning, and a valid observation can be formed only when both endpoints are detected in the current frame. Therefore, the candidate relation itself and the detection quality of the lower-confidence endpoint provide the most direct evidence in the current frame. The co-visibility term is weighted as $0.35$ because the two endpoints of a remote relation may be spatially distant and are not always stably visible at the same time. Co-visibility can strengthen the evidence, but should not be a decisive condition. The endpoint maturity term is also weighted as $0.35$, encouraging nodes with more stable geometry and more observations to participate in remote relations, while still allowing early-stage nodes to gradually accumulate support through subsequent multi-frame evidence. The distance prior is weighted as $0.45$, slightly higher than the co-visibility and maturity terms, because remote relations do not rely on local contact but should still satisfy basic spatial plausibility. Meanwhile, this weight is lower than the combined contribution of detection and confidence, avoiding excessive suppression of true remote relations caused by online pose errors or depth noise.

Geometric maturity is determined by the currently available geometric state of a node. Stable anchor-fused geometry has the highest maturity, followed by stable anchors, candidate anchors, and geometry estimated only in world coordinates. The distance prior is set according to the relation type: contact or installation relations prefer nearby endpoints, control or power-supply relations allow larger spatial separations, and other relations use a generic distance decay. This design makes remote relations more likely to select endpoint pairs with high detection confidence, mature geometry, current-frame co-visibility, and spatially plausible configurations.

For each label-level remote relation, the system maintains a temporal posterior over instance-level endpoint pairs. When a new remote observation arrives, its support is accumulated onto the corresponding source-target node pair, and different candidate endpoint pairs compete with each other. A remote relation edge is written to or refreshed in the functional graph only when one endpoint pair receives continuous support across multiple frames and remains stable over a recent time window. If the evidence is still insufficient, the system keeps the relation in a tentative state and does not directly commit it as a stable remote edge.

In the early stage of online mapping, one or both endpoints of a remote relation may not yet be stably associated with persistent nodes. To avoid losing such early observations, the system stores temporarily unresolved remote evidence in a bounded cache and re-attempts resolution after node association becomes more stable. If the cached evidence can be uniquely resolved to a source-target node pair, it is injected into the corresponding remote relation posterior. If multiple possible endpoint pairs still exist, the evidence remains unresolved and is not written into the functional graph.

In addition, the system uses scene-level remote relation templates as weak priors to recall possible remote relation candidates. These templates do not directly determine the final remote edges. Instead, a template-induced candidate is considered only after passing endpoint maturity, recent co-visibility, and spatial consistency checks, and is then accumulated through the same temporal posterior mechanism. Therefore, remote relations are committed only when they are consistently supported by instance-level endpoint association and multi-frame evidence, rather than by a single-frame label-level prediction or a template prior alone.

\subsection{Parameter Sensitivity Analysis}

Since the supplementary material involves multiple hyperparameters for node association, remote relations, and functional-topology-assisted loop closure, we further conduct a parameter sensitivity analysis on the FunGraph3D dataset. Specifically, we perturb the main hyperparameters appearing in the supplementary material by $\pm 20\%$ individually, while keeping all other settings unchanged, and report the maximum changes of each metric relative to the default configuration.

The experimental results show that Functional-SLAM is not sensitive to these hyperparameters. Across all perturbation experiments, the maximum variation of the average ATE RMSE is $5.3\%$, the maximum variation of Overall R@* in Node Evaluation is $6.8\%$, and the maximum variation of Overall R@* in Triplet Evaluation is $4.7\%$. These results indicate that the system performance does not rely on a finely tuned set of thresholds or weights, but remains stable over a relatively wide parameter range.

This robustness mainly comes from three aspects. First, node association does not depend on a single cue, but jointly considers geometric overlap, 3D distance, semantic labels, and functional context. Therefore, a small change in an individual weight does not directly determine the matching result. Second, functional edge commitment relies on multi-frame evidence and recent consistency rather than single-frame thresholding, making it tolerant to perturbations in relation confidence and support thresholds. Finally, functional-topology-assisted loop closure only generates supplementary candidates, while the final loop constraints still need to pass MASt3R geometric verification. Therefore, even if the thresholds related to topological candidates change, they do not directly introduce unverified false loop closures. Overall, the parameter sensitivity experiments show that the performance improvement of Functional-SLAM mainly comes from online functional graph maintenance and functional topology constraints themselves, rather than from over-reliance on specific hyperparameters.

\subsection{Baseline Adaptation and Relation Evaluation Protocol}

\subsubsection{Baseline Output Adaptation.}
Different functional scene graph methods use different graph representations and relation-text spaces. Directly using their raw outputs for node and triplet evaluation may therefore introduce additional representation bias. To ensure fair cross-method comparison, we conservatively adapt the outputs of several baseline methods only during evaluation. This adaptation does not modify the nodes, geometric positions, or edge connections predicted by the baseline methods, nor does it access GT relation labels. It only converts the structural edges or affordance texts of the baseline methods into node labels and functional relation descriptions that are more consistent with the OpenFunGraph benchmark protocol.

\paragraph{KeySG.}
KeySG adopts a hierarchical scene graph representation, where its edges mainly describe structural ownership between objects and functional parts, rather than interactive functional relations in the OpenFunGraph benchmark. For example, the edge text in KeySG is often of the form ``handle is a functional element of cabinet'' or ``knob is a functional element of stove''. Such descriptions indicate which object a functional part belongs to, but they cannot be directly aligned with benchmark relation texts such as ``pull to open or close'' or ``rotate to adjust the setting''. Therefore, directly using the original KeySG edge texts for relation evaluation would systematically underestimate its Edge Prediction recall, because the two methods use different relation semantic spaces.

To address this issue, we use an LLM to build a conservative relation-text adapter for KeySG. The adapter generates functional relation texts only from the functional part labels and parent-object labels predicted by KeySG itself. For example, a structural edge between handle and cabinet is mapped to ``pull to open or close'', a structural edge between knob and stove is mapped to ``rotate to adjust the setting'', and a structural edge between button and remote control is mapped to ``control, turn on or turn off''. This process does not change the nodes or edge endpoints predicted by KeySG. It only converts structural ownership descriptions into functional relation descriptions that can be evaluated by the benchmark, and therefore does not introduce additional oracle information.

\paragraph{FunGraph.}
The original FunGraph output contains a \texttt{small\_objects} field, which we use to distinguish functional parts from regular objects. Specifically, nodes appearing in \texttt{small\_objects} are regarded as interaction elements or functional parts, while nodes not appearing in \texttt{small\_objects} are regarded as objects. During evaluation, we only keep edges between small parts and objects. If both endpoints of an edge are small parts or both endpoints are objects, the edge is not used for OpenFunGraph-style triplet evaluation.

The original edges in FunGraph are also closer to structural or ownership relations, indicating that a small part belongs to an object. To compare FunGraph under the triplet recall definition of the OpenFunGraph benchmark, we convert its small-part--object structural edges into the endpoint form of ``a functional part acting on an object''. We emphasize that this step does not allow FunGraph to generate an additional functional graph. It only converts its original structural edges into an object--interaction-element relation form that can be evaluated by the benchmark.

In addition, the small-part labels in FunGraph often contain affordance descriptions, such as pull, rotate, and key press, whereas the node labels in the benchmark are usually entity names such as handle, knob, and button. Therefore, we apply deterministic label mapping to functional-part nodes: hook pull, pinch pull, and pull are mapped to handle; rotate and hook turn are mapped to knob; key press, tip push, and press are mapped to button; and foot push is mapped to pedal. Non-affordance original labels are kept unchanged.

For relation texts, we also generate the corresponding functional relation descriptions from the affordances predicted by FunGraph itself. Specifically, hook pull, pinch pull, and pull are mapped to ``pull to open or close''; rotate and hook turn are mapped to ``rotate to open or close''; turn is mapped to ``turn on or turn off''; push and foot push are mapped to ``push to open or close''; tip push, press, and key press are mapped to ``press to turn on or turn off''; and touch is mapped to ``control, turn on or turn off''. This mapping relies only on the affordances predicted by FunGraph and does not use GT relation labels.

\subsubsection{Relation Evaluation Protocol.}
The relation annotations in the OpenFunGraph benchmark are not discrete class IDs, but natural-language functional descriptions. Therefore, even when different methods predict the same functional relation, their outputs may fail to strictly hit the top-$K$ retrieval results from a fixed relation vocabulary because of differences in wording, object complements, or description granularity. For example, the predicted relation ``rotate to adjust the oven setting'' and the GT relation ``rotate to adjust setting or temperature'' both express the core semantics of adjusting the setting or temperature of an appliance by rotating an interaction part. However, the former explicitly includes the object ``oven'', while the latter uses the more general phrase ``setting or temperature''. Similarly, ``turn to control water flow and temperature'' and ``control the water flow'', as well as ``pull to open or close the cabinet door'' and ``pull to open or close a cabinet or drawer'', are semantically consistent in their core functions but differ in the granularity of natural-language expression.

If evaluation relies only on top-$K$ retrieval from a fixed vocabulary, such reasonable natural-language paraphrases may be incorrectly counted as relation prediction failures, thereby underestimating the functional relation prediction ability of a method. Therefore, while preserving the original relation top-$K$ retrieval rule of OpenFunGraph, we introduce BERT semantic similarity as an auxiliary criterion. Specifically, a relation is considered semantically matched if the GT relation appears in the top-$K$ retrieval results of the predicted relation, or if the BERT cosine similarity between the predicted relation and the GT relation is no lower than $0.70$.

This auxiliary rule is applied only to triplets whose endpoint nodes have already been successfully matched. In other words, it does not relax the requirements on object nodes, interaction-element nodes, 3D geometry, or node association. It is used only to reduce relation false negatives caused by differences in natural-language expression. Therefore, the protocol still imposes strict constraints on endpoint prediction and functional relation prediction, while more reasonably evaluating the semantic consistency of natural-language functional relation descriptions.

\subsection{Prompts}

To improve experimental reproducibility, we provide the main LLM prompts used in Functional-SLAM in the supplementary material. The prompts in the system are used only to generate structured functional observation priors from open-vocabulary labels, including the scene type, scene-level functional atlas, frame-level $O$-$C$-$U$ functional structures, and remote relation candidates. They do not directly access GT nodes, GT relations, or evaluation labels, and therefore do not introduce additional annotation information.

The prompt pipeline of Functional-SLAM consists of three levels. First, the system uses a scene classification prompt to infer the scene type from the open-vocabulary tag list detected in the current frame or sequence. The output is one of kitchen, bathroom, livingroom, bedroom, or unknown. This step uses the detected tags as evidence. The inferred scene type is then used to select the subsequent scene-level atlas prompt and incremental reasoning prompt.

Second, the system constructs a global functional atlas for common indoor scenes. The atlas prompt asks the LLM to generate scene-level manipulable objects $O$, functional carriers $C$, interaction units $U$, local functional chains, and remote functional relation candidates. Here, $O$ denotes scene objects, $C$ denotes intermediate functional structures inside objects that carry interactions, such as doors, drawers, or control panels, and $U$ denotes directly operable interaction units, such as handles, knobs, buttons, caps, or switches. The prompt explicitly distinguishes local relations from remote relations. Local relations are represented by the $O$-$C$-$U$ hierarchy or direct $O$-$U$ relations, while remote relations are used only to describe non-containment functional dependencies between different entities, such as a switch panel controlling a ceiling light or an electric outlet powering an appliance. The atlas prompt also asks the LLM to provide short functional relation descriptions for $O$-$C$, $C$-$U$, $O$-$U$, and remote edges, thereby providing structured priors for subsequent detection, node association, and relation posterior estimation.

Third, the system uses a frame-level incremental prompt to process the open-vocabulary tags in the current frame. This prompt receives the current scene type, the current-frame tag list, and a whitelist of possible remote-relation endpoints from the atlas. Its goal is to select a small number of specific, operable, and non-generic object categories from the current-frame tags, and to generate local $O$-$C$-$U$ structures, direct $O$-$U$ interaction units, and remote relation candidates for them. The prompt explicitly requires tag filtering and object selection before relation generation. Generic categories, such as ``device'', ``object'', and ``equipment'', are not allowed to serve as functional objects or remote endpoints. If the scene type is known, tags that clearly conflict with the scene are also excluded. This reduces long-tail noise and cross-scene false retrievals in open-vocabulary perception.

\end{document}